\documentclass[11pt]{article}

\usepackage[preprint]{acl}
\usepackage{times}
\usepackage{latexsym}

\usepackage[T1]{fontenc}

\usepackage[utf8]{inputenc}
\usepackage{algorithm}

\usepackage{microtype}

\usepackage{inconsolata}

\usepackage{graphicx}

\usepackage{hyperref}

\usepackage{amsmath}
\usepackage{amssymb}
\usepackage{booktabs}
\usepackage{multirow}
\usepackage[table]{xcolor}
\usepackage{graphicx}
\usepackage{arydshln}

\definecolor{Red}{rgb}{0.768, 0.054, 0.054}
\definecolor{Green}{rgb}{0,0.4,0.7}
\hypersetup{
    colorlinks=true,
    citecolor=teal,
    linkcolor=Red,
    urlcolor=Green,
}

\title{%
  \rule{\linewidth}{1.1pt}\\[4mm]
  Agent Memory Distillation:\\Empowering Small LLM Agents with Hierarchical Teacher Memory\\[2mm]
  \rule{\linewidth}{1.1pt}%
}

\author{
  Taeil Kim$^{1*}$ \quad
  Kangsan Kim$^{1*}$ \quad
  Sung Ju Hwang$^{1,2}$ \\[4pt]
  $^{1}$KAIST \quad $^{2}$DeepAuto.ai \\[4pt]
  {\fontsize{10}{11}\selectfont \url{https://agent-memory-distillation.github.io/}} \\
  \small{\texttt{\{kti5589, kangsan.kim, sungju.hwang\}@kaist.ac.kr}}
}
\vspace{0.3in}

\begin{document}
\maketitle
\def\thefootnote{*}\footnotetext{Equal contribution}\def\thefootnote{\arabic{footnote}}

\begin{abstract}
Memory systems have shown promise for improving agent performance, but their potential 
remains largely unexplored for small language models, which struggle to generate 
sufficient successful trajectories on their own.
We propose \textbf{Agent Memory Distillation (AMD)}, a training-free framework that 
transfers structured knowledge from a large teacher agent to a small student agent 
through hierarchical memory.
AMD constructs three complementary memory types from successful teacher trajectories: 
\texttt{Workflow} memory encodes task-level strategies, \texttt{Subtask} memory provides 
concrete behavioral examples at an intermediate granularity, and \texttt{Function} memory 
captures per-function calling conventions and common pitfalls.
\texttt{Workflow} and \texttt{Subtask} memories are injected proactively at the start of 
each task, while \texttt{Function} memory is retrieved reactively upon tool-calling errors.
We evaluate AMD on three tool-use benchmarks using four student models (4B--8B parameters) 
with GPT-5-mini as the teacher, achieving average accuracy gains of 27.2\%p, 11.2\%p, and 
3.4\%p on AppWorld, BFCL V3, and ToolSandbox, while consistently outperforming existing 
memory-based baselines.
Further analysis shows that \texttt{Subtask} memory contributes the largest gains, teacher effectiveness depends on both teacher capability and student compatibility, and 4B-sized students benefit most from AMD.
\end{abstract}

\section{Introduction}

Memory has emerged as a critical component for developing capable self-evolving agents, enabling the reuse of successful behavioral patterns and the avoidance of past failures~\cite{zhao2024expel, ouyang2025reasoningbank, zhang2026memrl}.
As modern agents increasingly rely on external tools such as APIs and predefined functions to tackle complex tasks, the role of memory becomes even more essential~\cite{liao2025reflectool, xu2026evolution}.
In such settings, memory helps agents recall effective tool-use strategies from past interactions, improving both efficiency and task success rates~\cite{fang2025memp, xia2025experience}.
Moreover, memory can encode tool conventions such as argument schemas and return structures, enabling agents to invoke tools more accurately and reliably~\cite{du2026memory}.

\begin{figure}[t!]
    \centering
    \includegraphics[width=\linewidth]{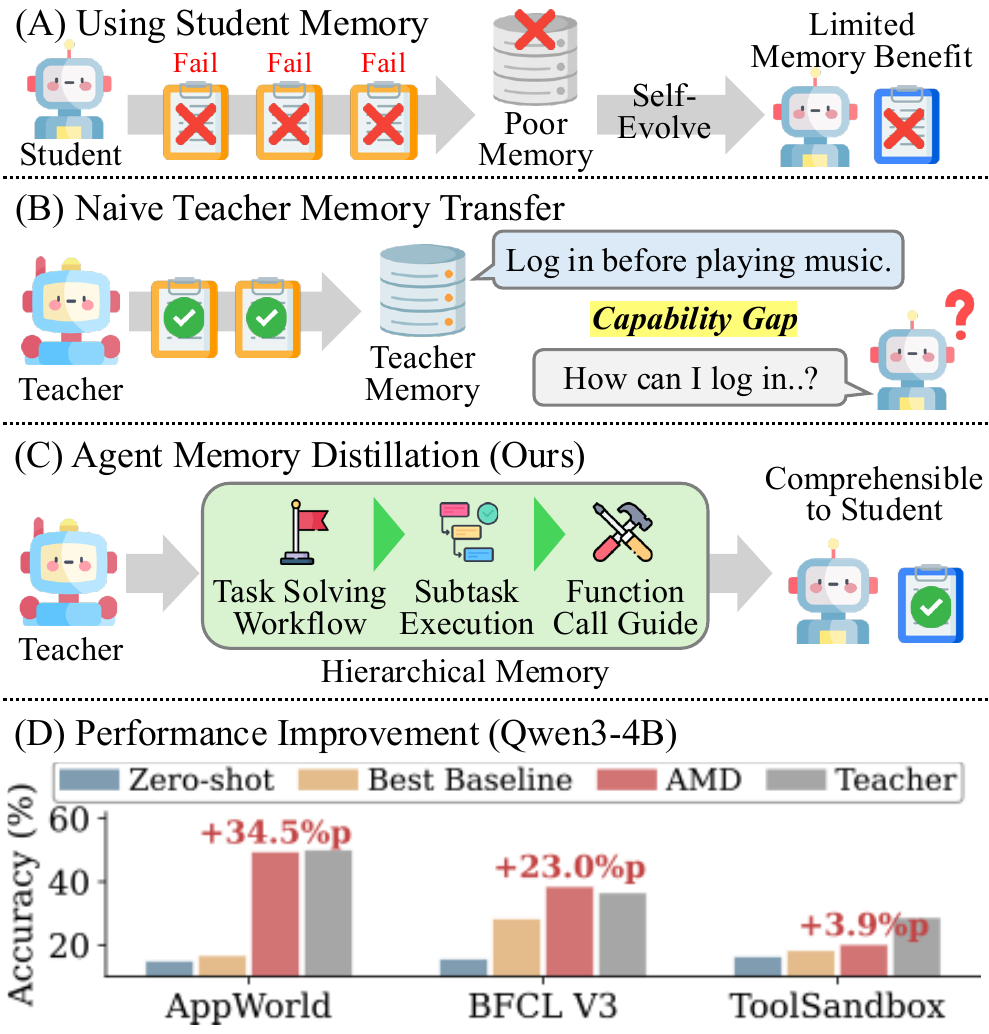}
    \vspace{-0.27in}
    \caption{\textbf{Motivation, Concept, and Results of AMD.} 
(A) Student-generated memory is limited by low task success rates. 
(B) Naive teacher memory transfer yields marginal gains due to the capability gap. 
(C) AMD transfers hierarchical memories spanning task, subtask, and function levels, making teacher knowledge 
accessible to small students. (D) AMD achieves significant accuracy gains across three benchmarks.}
    \label{fig:concept}
    \vspace{-0.1in}
\end{figure}

However, the potential of memory remains largely unexplored for small agents, in contrast to its demonstrated effectiveness with large proprietary models~\cite{wu2025human, luo2026storage}.
A key challenge is that small agents typically exhibit lower task success rates, resulting in memory repositories that are dominated by unsuccessful trajectories and contain only a limited number of successful experiences, as shown in \autoref{fig:concept} (A).
Although agents can reflect on past failures to improve subsequent actions, the scarcity of successful trajectories fundamentally limits the impact of memory utilization~\cite{hu2025sample, allard2026experiential, ding2026agenther}.
To overcome this limitation, we propose leveraging memories generated by stronger teacher agents, whose superior performance yields a rich source of successful trajectories and tool-use examples. A related line of work also transfers teacher experience to small students, for example by extracting hints from failed trajectories to generate successful teacher trajectories and then fine-tuning the student on them~\cite{ibrahim2025finetuning}.
Inspired by knowledge distillation~\cite{hinton2015distilling, kang2026distilling}, our method enhances student agents in a training-free manner by directly transferring the teacher's experiences rather than optimizing model parameters.

Despite access to high-quality teacher experience, we observe that naive memory transfer yields only marginal improvements, comparable to those achieved by student-generated memory alone.
This suggests that the capability gap between teacher and student agents remains a major obstacle to effective memory distillation, analogous to the knowledge gap observed in conventional knowledge distillation~\cite{mirzadeh2020improved, guo2020reducing}.
For example, a teacher memory may recommend a high-level strategy such as \textit{"Log in first to start the playlist,"} while the student lacks the prerequisite knowledge to execute the login procedure itself, as shown in \autoref{fig:concept} (B).
Furthermore, small models are known to exhibit limited in-context learning ability and weaker instruction-following capacity compared to larger models, leading to difficulties in effectively interpreting and applying teacher memories even when they are directly provided~\cite{wei2022emergent, zhao2024context, shen2024small}.
These factors collectively highlight the need for a principled knowledge transfer strategy tailored to agent memory distillation.

In this work, we introduce \textbf{Agent Memory Distillation (AMD)}, a novel framework for transferring teacher agent experiences to small student agents. 
Rather than relying on a single memory representation, AMD constructs three types of memory from teacher experiences, organized in a hierarchical structure spanning task-level, subtask-level, and function-level granularities to facilitate effective comprehension and 
application of teacher knowledge, as illustrated in \autoref{fig:concept} (C).
Specifically, \texttt{Workflow} memory captures overall task completion strategies, enabling the student to decompose tasks into subtasks and establish a coherent high-level plan. \texttt{Subtask} memory provides concrete action examples for each subtask, allowing the student to reference successful teacher behaviors at an intermediate level. Finally, \texttt{Function} memory encodes detailed tool schemas and usage examples, which are retrieved when the student encounters tool-calling errors during inference to guide correct tool execution.

To validate the effectiveness of AMD, we conduct experiments across three benchmarks using four small language models (4B or 8B parameters) as student agents, with GPT-5-mini serving as the teacher.
AMD consistently achieves substantial performance improvements over the zero-shot baseline across all student models, with \textbf{average accuracy gains of 27.2\%p on AppWorld}~\cite{trivedi2024appworld}, \textbf{11.2\%p on BFCL V3}~\cite{patil2025bfcl}, and \textbf{3.4\%p on ToolSandbox}~\cite{lu2024toolsandbox}, while outperforming all 
memory-based baselines that are not designed for the memory distillation setting~\cite{ouyang2025reasoningbank, fang2025memp, shen2026structurally}.
Beyond accuracy, AMD also reduces the number of interaction turns required by the student, yielding trajectories that more closely resemble those of the teacher agent. We further reveal that each memory type contributes distinctly to knowledge transfer, that transfer quality depends on both teacher accuracy and teacher-student compatibility, and that 4B-scale students tend to benefit most from teacher memory distillation.

In conclusion, this work presents the first systematic investigation into effective teacher-to-student memory transfer for small agents. By first identifying the limitations of naive memory transfer, we propose a hierarchically structured memory distillation framework that enables small student agents to better comprehend and apply teacher knowledge across multiple levels of task granularity. 
We hope this work provides a solid foundation for future research on agent memory distillation and the development of more capable small LLM agents by enhancing their ability to effectively leverage the accumulated experiences of stronger agents.
\section{Related Work}

\subsection{LLM Agents with Memory}
Memory enables agents to leverage previous experiences by following successful patterns while avoiding repeated failures~\cite{zheng2024synapse, zhang2025memevolve, tang2025agent, zhang2026memrl}. Reflexion~\cite{shinn2023reflexion} first demonstrates the potential of verbalized self-feedback and episodic memory buffers for agent self-improvement.  
ExpeL~\cite{zhao2024expel} enables agents to collect experience, extract cross-task insights, and retrieve them during future inferences.
Similarly, AWM~\cite{wang2024agent} extracts and reuses common multi-step workflows, while ReasoningBank~\cite{ouyang2025reasoningbank} samples candidate trajectories through test-time scaling and generates reasoning insights for upcoming tasks. 
MemP~\cite{fang2025memp} proposes procedural memory that distills agent trajectories into multi-level abstractions and investigates effective memory management mechanisms.
ReflecTool~\cite{liao2025reflectool} stores tool-wise experience in memory and retrieves relevant trajectories at inference time. 
SASM~\cite{shen2026structurally} further proposes subtask-level memory alignment to address granularity mismatch in instance-level memory retrieval.
However, existing methods are predominantly evaluated with large proprietary models such as GPT-4~\cite{achiam2023gpt} or Gemini 2.5~\cite{comanici2025gemini}, overlooking the limitation of memory utilization in small agents stemming from their degraded reasoning and instruction-following capabilities. 
We address this gap by providing small agents with structured, multi-level knowledge distilled from teachers in a detailed and accessible form.

\subsection{Knowledge Distillation in LLM}
Knowledge distillation~\cite{hinton2015distilling} has long been studied as a paradigm for transferring knowledge from a stronger teacher model to a smaller student model, with demonstrated effectiveness across a wide range of LLM settings~\cite{sanh2019distilbert, hsieh2023distilling, gu2024minillm, guo2025deepseek}. 
Recent works have extended this paradigm to agentic settings. 
SAD~\cite{liu2025structured} segments teacher trajectories into reasoning and action spans and applies segment-wise distillation losses. 
Agent Distillation~\cite{kang2026distilling} transfers full task-solving behaviors, including retrieval and code tool use, from teacher to student models, and SCoRe~\cite{lyu2025correction} introduces a reinforced distillation framework based on student-centered short-horizon reinforcement learning. 
However, these approaches rely on parameter updates through costly training 
and do not explore memory-based knowledge transfer. AgentDistill~\cite{qiu2025agentdistill} takes a training-free approach by directly reusing teacher-generated MCPs, yet it does not enable students to leverage teacher memory directly. 
Memory Transfer Learning~\cite{kim2026memory} investigates inter-model knowledge transfer but is not designed for the teacher-student distillation setting. 
It is also well established that the capability gap between teacher and student models can fundamentally limit distillation effectiveness~\cite{mirzadeh2020improved, guo2020reducing, xu2025speculative}.
To the best of our knowledge, AMD is the first work to systematically address teacher-to-student memory distillation, explicitly accounting for the capacity gap through a principled hierarchical memory transfer design.
\section{Agent Memory Distillation}

\begin{figure*}[t!]
    \centering
    \vspace{-0.1in}
    \includegraphics[width=\textwidth]{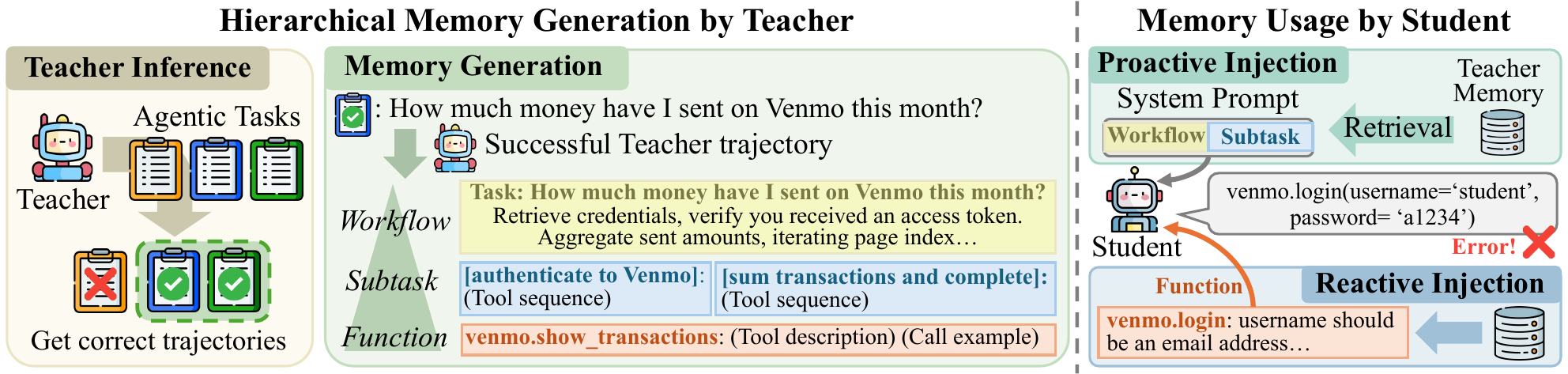}
    \vspace{-0.25in}
    \caption{\textbf{Hierarchical Memory Generation and Injection in AMD.} 
(\textit{Left}) The teacher agent generates three types of 
memory from successful trajectories. (\textit{Right}) At inference time, \texttt{Workflow} and \texttt{Subtask} memories are proactively injected into the system prompt, while \texttt{Function} memory is reactively retrieved upon tool-calling errors.}
    \label{fig:method}
    \vspace{-0.2in}
\end{figure*}

We present AMD, a novel framework that transfers teacher memory to a small student agent.
We first formalize the problem setting in \autoref{sec:problem_formulation}, then describe multi-level memory generation by a teacher in \autoref{sec:memory_generation} and memory retrieval and utilization by a student in \autoref{sec:memory_retrieval}.

\subsection{Problem Formulation}
\label{sec:problem_formulation}

We focus on multi-turn tool-use reasoning tasks, where an agent $\pi$ operates in an interactive environment by issuing a sequence of tool calls from a predefined set $\mathcal{F}$ and receiving observations in return.
Given a task $s$, the agent produces a trajectory $\tau = \bigl((a_1, o_1), \ldots, (a_T, o_T)\bigr)$, where $a_t \in \mathcal{F}$ is a tool call and $o_t$ is the resulting observation at step $t$.
We distinguish between a teacher agent $\pi^T$, backed by a large and capable language model, and a student agent $\pi^S$, backed by a small model (4B or 8B parameters).
Given a task set $\mathcal{S}$, we first run $\pi^T$ on $\mathcal{S}$ to collect a set of teacher trajectories $\mathcal{D}^T = \{\tau^T_1, ..., \tau^T_N\}$, where $N = |\mathcal{S}|$. 
We then use $\pi^T$ to construct a memory store $\mathcal{M}$ from $\mathcal{D}^T$, which is subsequently transferred to $\pi^S$.
At inference time, $\pi^S$ retrieves relevant memories from $\mathcal{M}$ to guide its reasoning on $\mathcal{S}$. 
The goal of AMD is to maximize the task performance of $\pi^S$ through effective utilization of teacher-generated memories:
\begin{equation}
    \max_{\mathcal{M}} \;
    \mathbb{E}_{s \sim \mathcal{S}}
    \left[R\!\left(\pi^S(s;\mathcal{M})\right)\right],
\end{equation}
where $R(\pi^S(s; \mathcal{M}))$ denotes the task success reward obtained by $\pi^S$ on task $s$ given memory $\mathcal{M}$.

\subsection{Hierarchical Memory Generation}
\label{sec:memory_generation}

AMD constructs three types of memory from the teacher's past experiences, organized at different levels of task granularity.
AMD employs $\pi^T$ to construct each memory type from $\mathcal{D}^T_{+} \subseteq \mathcal{D}^T$, 
the subset of successful teacher trajectories, producing a memory store 
$\mathcal{M} = \mathcal{M}^{wf} \cup \mathcal{M}^{st} \cup \mathcal{M}^{fn}$, where $\mathcal{M}^{wf}, \mathcal{M}^{st}, \text{and } \mathcal{M}^{fn}$ denote \texttt{Workflow}, \texttt{Subtask}, and \texttt{Function} memories.



\paragraph{Workflow Memory}

\texttt{Workflow} memory captures the teacher's high-level task-completion strategy.
For each successful trajectory $\tau^T_i \in \mathcal{D}^T_{+}$, a verbalized insight is produced that describes the overall approach taken by the teacher in natural language, covering the apps and tools involved, key preconditions, and decision rules for task completion, as well as validation cues and common failure patterns to avoid.
The insight abstracts over concrete runtime values: identifiers, credentials, file paths, and other dynamic inputs are replaced with typed placeholders (e.g., \texttt{<ID>}, \texttt{<EMAIL>}, \texttt{<FILE\_PATH>}), so that the memory remains applicable to future tasks with different specific inputs.
Together with a natural language query that characterizes the task, the insight forms a workflow memory entry $m^{wf}_i = (q_i,\, \text{ins}_i)$, which is encoded into a dense vector for retrieval.
The resulting memory bank $\mathcal{M}^{wf} = \{m^{wf}_i\}_{i=1}^{|\mathcal{D}^T_{+}|}$ provides the student agent with a high-level task plan before execution begins, enabling it to identify the relevant tools and establish a coherent action sequence.

\paragraph{Subtask Memory}

\texttt{Subtask} memory provides concrete behavioral examples at an intermediate level of granularity, bridging the gap between high-level workflow plans and low-level tool calls.
For each successful trajectory $\tau^T_i \in \mathcal{D}^T_{+}$, we decompose the trajectory into a sequence of coherent subtask segments $\{e_{i,1}, \ldots, e_{i,K_i}\}$, where each segment corresponds to a semantically meaningful unit of the teacher's behavior, such as authenticating with a service or executing a sequence of related API calls to fulfill one subtask.
Segmentation is performed by a teacher LLM prompted to identify semantically coherent units in the trajectory, optionally guided by rule-based heuristics that provide candidate breakpoint hints.
Each segment $e_{i,k}$, which contains the concrete execution examples (tool calls or executable code paired with the corresponding observations from the teacher), is stored together with a label $\ell_{i,k}$, a short natural language description $d_{i,k}$, forming a subtask memory entry $m^{st}_{i,k} = (\ell_{i,k},\, d_{i,k},\, e_{i,k})$.
Each description $d_{i,k}$ is encoded into a dense vector for retrieval, and all
entries across all trajectories constitute the subtask memory bank
$\mathcal{M}^{st} = \bigcup_{i=1}^{|\mathcal{D}^T_{+}|}
\{m^{st}_{i,k}\}_{k=1}^{K_i}$.



\paragraph{Function Memory}
\texttt{Function} memory captures fine-grained tool invocation knowledge at the level of individual function calls, built directly from the successful teacher trajectories and optionally augmented with the corresponding function documentation.
For each successful trajectory $\tau^T_i \in \mathcal{D}^T_{+}$, we extract its constituent function invocations, indexed by $j \in \{1, \ldots, J_i\}$.
Each record stores the function name $f_{i,j}$ and a concrete teacher example $E_{i,j}$, and is optionally augmented with the API documentation $\text{doc}(f_{i,j})$ that specifies the argument and response schema.
The example $E_{i,j}$ comprises the executable invocation together with its surrounding context, along with any returned observation when available.
This context makes the rationale for the call apparent, namely why the function was invoked at that step and what constraints govern its arguments.
Together these form a function memory entry $m^{fn}_{i,j} = (f_{i,j},\, E_{i,j},\, \text{doc}(f_{i,j}))$.
Unlike the workflow and subtask banks queried through dense vector similarity, function entries are indexed by function name $f_{i,j}$, constituting the memory bank $\mathcal{M}^{fn} = \bigcup_{i=1}^{|\mathcal{D}^T_{+}|}\{m^{fn}_{i,j}\}_{j=1}^{J_i}$.

\subsection{Memory Retrieval and Injection}
\label{sec:memory_retrieval}
At inference time, $\pi^S$ retrieves relevant memories from $\mathcal{M}$ via embedding-based cosine similarity, where all memory entries are pre-encoded using a pretrained text embedding model. 
Memories below a minimum similarity threshold are discarded. 

\paragraph{Proactive Injection}
\texttt{Workflow} and \texttt{Subtask} memories are injected once at the beginning of each task, before the agent begins execution.
For \texttt{Workflow} memory, the task instruction serves as the retrieval query, and the top-$k$ entries are retrieved from $\mathcal{M}^{wf}$. 
The retrieved insight is prepended to the system prompt, providing $\pi^S$ with a high-level plan before it issues any tool calls.

For \texttt{Subtask} memory, a student $\pi^S$ first decomposes the task instruction into an ordered sequence of subtask labels (up to six). 
Each subtask label is used as an independent retrieval query against $\mathcal{M}^{st}$, and the best-matching memory $m^{st}$ is retrieved per subtask with deduplication across subtasks to avoid redundant examples. 
The retrieved segments, including their execution examples and observations, are injected alongside \texttt{Workflow} in the system prompt.


\paragraph{Reactive Injection}

\texttt{Function} memory is retrieved reactively in response to execution failures. 
When a tool call returns an error, the name of the failing function is used to look up candidate records in $\mathcal{M}^{fn}$, which is indexed by function name. 
When multiple records exist for the same function, they are ranked by the cosine similarity between the current task instruction and the stored reasoning of each record, and the top examples are selected. 
The retrieved records are formatted as a hint block and appended to the error message in the current context, providing $\pi^S$ with targeted corrective guidance at the moment of failure without inflating the context during successful execution.
\section{Experiment}

\subsection{Experimental Setup}


\paragraph{Benchmarks}
We evaluate AMD on three benchmarks that cover diverse tool-use settings.
\textbf{AppWorld}~\cite{trivedi2024appworld} is a multi-app agent benchmark in which an agent must complete complex, multi-step tasks by interacting with a set of simulated real-world applications such as email, messaging, and payment services via Python API calls.
The unit of action is a block of Python code executed in a stateful interpreter, and success is measured by database-state unit tests.
We evaluate on the \texttt{test\_normal} subset consisting of 168 tasks.
\textbf{BFCL V3}~\cite{patil2025bfcl} is a function-calling benchmark that evaluates an agent's ability to invoke the correct functions with accurate arguments.
Each action of the agent is a structured function call, and user turns are pre-specified.
An entry is correct only if every turn matches the expected API state and a minimal viable call path. 
We use the multi-turn base subset of 200 tasks.
\textbf{ToolSandbox}~\cite{lu2024toolsandbox} is a stateful, conversational tool-use benchmark where tools depend on a shared world state and on prior tool calls.
Each action of the agent is a JSON tool call, with an LLM-simulated user driving the dialogue.
Trajectories are scored against human-authored milestones and minefields.
We evaluate on the base subset of 129 scenarios, using GPT-5-mini as the user simulator.

\paragraph{Baselines}
We compare AMD against the teacher agent and student \textbf{Zero-shot} performance, as well as three representative agent memory frameworks adapted to the teacher-to-student transfer setting.
\textbf{ReasoningBank}~\cite{ouyang2025reasoningbank} represents a flat, task-level memory approach that retrieves a single reasoning insight per task.
\textbf{MemP}~\cite{fang2025memp} organizes memory into hierarchical procedural abstractions, but is designed for same-model self-evolution rather than cross-model transfer.
\textbf{SASM}~\cite{shen2026structurally} introduces subtask-level retrieval granularity, but applies a single uniform memory type without distinguishing between planning, execution, and error-recovery knowledge.
Memory is generated from the same set of teacher trajectories and applied at student inference time following each method's proposed protocol.

\paragraph{Implementation Details}
We use GPT-5-mini as the teacher agent and evaluate four student models: Qwen3-4B, Qwen3-8B~\cite{yang2025qwen3}, Gemma4-E4B~\cite{gemma4_e4b_2026}, and Llama3.1-8B~\cite{grattafiori2024llama}.
Memory entries are encoded using OpenAI's \texttt{text-embedding-3-small} model.
At inference time, we retrieve the top-1 \texttt{Workflow} memory entry and the top-1 \texttt{Subtask} segment per decomposed subtask, and the top-1 \texttt{Function} memory record per failing function call ($k=1$).
To ensure robust evaluation, each experiment was repeated twice, and their average performance is reported.

\begin{table}[t]
\centering
\vspace{-0.1in}
\caption{\textbf{Main results across three benchmarks.} $\Delta$ shows absolute accuracy gains of AMD over zero-shot.}
\label{tab:main_acc}
\vspace{-0.1in}
\setlength{\tabcolsep}{4pt}
\renewcommand{\arraystretch}{0.85}
\resizebox{\linewidth}{!}{%
\begin{tabular}{l cccc}
\toprule
\textbf{Method}
  & \textbf{AppWorld}
  & \textbf{BFCL V3}
  & \textbf{ToolSandbox}
  & \textbf{Average} \\
\midrule
\midrule
\rowcolor{green!10}
\multicolumn{5}{l}{\textbf{\textit{Teacher Agent}}} \\
GPT-5-mini        & 50.00 & 36.50 & 28.68 & 38.39 \\
\midrule
\rowcolor{green!10}
\multicolumn{5}{l}{\textbf{\textit{Student Agents}}} \\
\rowcolor{gray!15}
\multicolumn{5}{l}{\textit{Qwen3-4B}} \\
\hspace{1em}Zero-shot      & 14.88 & 15.50 & 16.28 & 15.55 \\
\hspace{1em}ReasoningBank & 10.71 & 24.25 & 16.28 & 17.08 \\
\hspace{1em}MemP          & 16.67 & 28.25 & 14.73 & 19.88 \\
\hspace{1em}SASM           & 15.48 & 15.25 & 18.22 & 16.32 \\
\hspace{1em}\textbf{AMD}   & \textbf{49.40} & \textbf{38.50} & \textbf{20.16} & \textbf{36.02} \\
\rowcolor{blue!10}
\hspace{1em}\textbf{\textit{$\Delta$}} & \textbf{+34.52} & \textbf{+23.00} & \textbf{+3.88} & \textbf{+20.47} \\
\noalign{\vskip 0.25ex}\cdashline{1-5}\noalign{\vskip 0.75ex}
\rowcolor{gray!15}
\multicolumn{5}{l}{\textit{Gemma4-E4B}} \\
\hspace{1em}Zero-shot      & 24.40 & 37.25 & 18.22 & 26.62 \\
\hspace{1em}ReasoningBank & 30.36 & 40.25 & 15.12 & 28.58 \\
\hspace{1em}MemP          & 20.83 & 35.25 & 15.89 & 23.99 \\
\hspace{1em}SASM           & 22.02 & 36.75 & 12.40 & 23.72 \\
\hspace{1em}\textbf{AMD}   & \textbf{54.17} & \textbf{46.00} & \textbf{21.71} & \textbf{40.63} \\
\rowcolor{blue!10}
\hspace{1em}\textbf{\textit{$\Delta$}} & \textbf{+29.77} & \textbf{+8.75} & \textbf{+3.49} & \textbf{+14.00} \\
\noalign{\vskip 0.25ex}\cdashline{1-5}\noalign{\vskip 0.75ex}
\rowcolor{gray!15}
\multicolumn{5}{l}{\textit{Qwen3-8B}} \\
\hspace{1em}Zero-shot    & 25.60 & 38.00 & 20.16 & 27.92 \\
\hspace{1em}\textbf{AMD} & \textbf{51.79} & \textbf{45.50} & \textbf{25.58} & \textbf{40.96} \\
\rowcolor{blue!10}
\hspace{1em}\textbf{\textit{$\Delta$}} & \textbf{+26.19} & \textbf{+7.50} & \textbf{+5.42} & \textbf{+13.04} \\
\noalign{\vskip 0.25ex}\cdashline{1-5}\noalign{\vskip 0.75ex}
\rowcolor{gray!15}
\multicolumn{5}{l}{\textit{Llama3.1-8B}} \\
\hspace{1em}Zero-shot    &  8.93 &  9.00 &  5.43 &  7.79 \\
\hspace{1em}\textbf{AMD} & \textbf{27.38} & \textbf{14.50} & \textbf{6.20} & \textbf{16.03} \\
\rowcolor{blue!10}
\hspace{1em}\textbf{\textit{$\Delta$}} & \textbf{+18.45} & \textbf{+5.50} & \textbf{+0.77} & \textbf{+8.24} \\
\bottomrule
\end{tabular}%
}
\end{table}

\subsection{Experimental Results}
We report the evaluation results across all benchmarks and student models in \autoref{tab:main_acc}. 
AMD consistently outperforms all baselines in every model and benchmark, highlighting the effectiveness of hierarchical memory distillation from a teacher agent.

\paragraph{AMD vs. Baselines}
AMD achieves average accuracy gains of 27.2\%p, 11.2\%p, and 3.4\%p over zero-shot across the four student models on AppWorld, BFCL V3, and ToolSandbox, respectively.
In contrast, the three baseline memory methods yield inconsistent improvements and, in several cases, degrade performance relative to zero-shot.
For instance, ReasoningBank reduces Qwen3-4B accuracy on AppWorld from 14.88\% to 10.71\%, while MemP and SASM show similarly unstable behavior across models. These results suggest that directly transferring flat or inadequately structured teacher memory may introduce noise or exceed the comprehension and instruction-following capabilities of smaller students. 
AMD mitigates these limitations by organizing teacher knowledge hierarchically across multiple granularities and using representations appropriate for each memory level, making the transferred knowledge more accessible to the student agent.

\begin{figure}[t]
    \centering
    \vspace{-0.1in}
    \includegraphics[width=\linewidth]{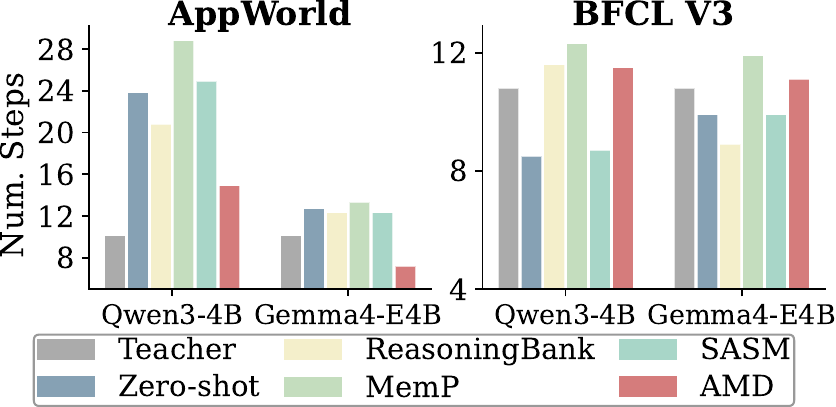}
    \vspace{-0.25in}
    \caption{\textbf{Interaction steps on two benchmarks.} AMD 
brings the student's turn count closer to the teacher's, 
particularly where the zero-shot gap is large.}
    \label{fig:step_fig}
    \vspace{-0.2in}
\end{figure}

\paragraph{Matching and Surpassing the Teacher}
AMD enables small student models to achieve teacher-level performance on several benchmarks.
On AppWorld, Gemma4-E4B (54.17\%) and Qwen3-8B (51.79\%) surpass GPT-5-mini performance (50.00\%) with AMD while Qwen3-4B shows comparable performance (49.40\%) with the teacher.
Moreover, on BFCL V3, three students outperform the teacher, and this advantage persists in aggregate: Gemma4-E4B (40.63\%) and Qwen3-8B (40.96\%) exceed the teacher's average accuracy (38.39\%).
Together, these results indicate that the student is not merely imitating the teacher's trajectories, instead, AMD distills transferable decision-making patterns that each student re-instantiates under its own inductive biases, allowing the distilled behavior to exceed the source of supervision.


\paragraph{Interaction Efficiency}
Beyond accuracy, AMD also shapes the student's interaction efficiency in a manner consistent with the teacher.
As shown in \autoref{fig:step_fig}, on AppWorld, zero-shot students issue far more turns than the teacher (e.g., 23.8 vs.\ 10.1 for Qwen3-4B), reflecting inefficient exploration without prior knowledge.
AMD substantially reduces this gap, bringing the student's turn count much closer to the teacher's (14.9 for Qwen3-4B, 7.2 for Gemma4-E4B).
On BFCL V3, where the zero-shot turn counts already align closely with the teacher's (around 8 to 11 turns), AMD preserves this efficiency without significant change.
This pattern suggests that AMD transfers not only what to do but also how efficiently to do it, and the degree of alignment tracks the original gap between teacher and zero-shot student behavior.

\section{Analysis}

\subsection{Ablation on Memory Components}

\begin{table*}[t]
\centering
\vspace{-0.1in}
\caption{\textbf{Ablation study across models and benchmarks.} WF, ST, and FN
denote workflow, subtask, and function memory.}
\label{tab:method_ablation}
\vspace{-0.1in}
\setlength{\tabcolsep}{2pt}
\renewcommand{\arraystretch}{0.9}
\resizebox{\textwidth}{!}{%
\begin{tabular}{l cccc cccc}
\toprule
\multirow{2}{*}{\textbf{Method}}
  & \multicolumn{4}{c}{\textbf{AppWorld}}
  & \multicolumn{4}{c}{\textbf{BFCL V3}} \\
\cmidrule(lr){2-5}\cmidrule(lr){6-9}
 & Qwen3-4B & Qwen3-8B & LLaMA3.1-8B & Gemma4-E4B
 & Qwen3-4B & Qwen3-8B & LLaMA3.1-8B & Gemma4-E4B \\
\midrule
\midrule
Zero-shot       & 14.88 & 25.60 &  8.93 & 24.40 & 15.50 & 38.00 &  9.00 & 37.25 \\
\midrule
WF              & 22.02 & 30.36 & 11.31 & 30.36 & 35.50 & 40.00 & 11.50 & 45.50 \\
WF + FN         & 24.11 & 33.93 & 14.88 & 40.48 & 35.50 & 41.50 & 12.50 & 46.00 \\
WF + ST         & 47.02 & 51.19 & \textbf{30.36} & 53.57 & 37.50 & 45.50 & 14.00 & 45.00 \\
WF + ST + FN    & \textbf{49.40} & \textbf{51.79} & 27.38 & \textbf{54.17} & \textbf{38.50} & \textbf{45.50} & \textbf{14.50} & \textbf{46.00} \\
\midrule
Student Memory  & 16.07 & 29.76 &  8.93 & 25.60 & 27.00 & 43.00 &  9.50 & 44.50 \\
\bottomrule
\end{tabular}%
}
\vspace{-0.1in}
\end{table*}

We report the contribution of each memory component by incrementally adding \texttt{Workflow}, \texttt{Subtask}, and \texttt{Function} memories on top of each other in \autoref{tab:method_ablation}.
\texttt{Workflow} memory alone yields consistent gains over zero-shot across both benchmarks, providing the student with a task-level plan before execution.
Adding \texttt{Subtask} memory produces the largest incremental improvement, particularly on AppWorld (e.g., +25.0\%p for Qwen3-4B over WF alone), confirming that concrete behavioral examples at the subtask level are the most critical component for complex, long-horizon tasks.
\texttt{Function} memory contributes additional gains when combined with \texttt{Workflow} and \texttt{Subtask}, though its effect is smaller in magnitude.
We also compare against a variant that uses student-generated memories in place of teacher memories (\textit{Student Memory}), which yields results close to zero-shot performance on AppWorld and substantially below AMD on both benchmarks.
This confirms that the quality and completeness of the teacher's trajectories are essential: the student's own experience, being less reliable, does not provide the structured knowledge necessary for effective memory-guided execution.

\begin{table}[t]
\centering
\vspace{-0.1in}
\caption{\textbf{Effect of teacher agents on AppWorld.} GPT-5-mini shows the strongest transfer effectiveness for Qwen3-4B student.}
\label{tab:teacher_ablation}
\vspace{-0.1in}
\resizebox{\linewidth}{!}{%
\begin{tabular}{l c cc}
\toprule
\textbf{Teacher} & \textbf{Teacher Acc} & \textbf{Qwen3-4B} & \textbf{Qwen3-8B} \\
\midrule
\midrule
Zero-shot        & -     & 14.88 & 25.60 \\
\midrule
GPT-5.5          & 91.08 & \underline{47.02} & \textbf{58.93} \\
DeepSeek V4 Pro  & 81.55 & 38.10             & \underline{57.14} \\
GPT-5-mini       & 50.00 & \textbf{49.40}    & 51.79 \\
Qwen3-32B        & 34.42 & 29.76             & 39.29 \\
\bottomrule
\end{tabular}%
}
\vspace{-0.1in}
\end{table}

\subsection{Effect of Teacher Agent}

We report the effect of substituting different teacher agents on AppWorld in \autoref{tab:teacher_ablation}.
For the stronger Qwen3-8B student, teacher accuracy is a reliable predictor of distillation quality: GPT-5.5 (91.08\%) yields the highest student accuracy (58.93\%), followed by DeepSeek V4 Pro (81.55\%, 57.14\%), GPT-5-mini (50.00\%, 51.79\%), and Qwen3-32B (34.42\%, 39.29\%).
A more accurate teacher completes more tasks successfully, supplying a larger and higher-quality pool of successful trajectories from which richer memories are extracted.
For the weaker Qwen3-4B student, however, this ordering breaks down: GPT-5-mini achieves the best student performance (49.40\%), outperforming even DeepSeek V4 Pro (38.10\%) despite its substantially lower teacher accuracy.
These results indicate that AMD substantially mitigates the capability gap, although a modest gap remains between the 4B student and the teacher models, suggesting room for further refinement of memory distillation.

\subsection{Effect of Student Model Size}
\begin{figure}[t!]
    \centering
    \vspace{-0.1in}
    \includegraphics[width=\linewidth]{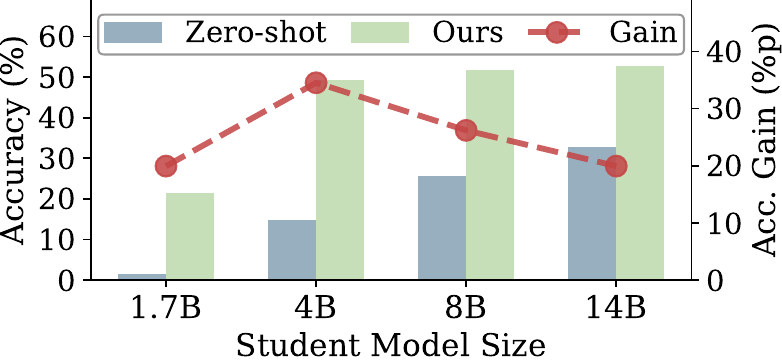}
    \vspace{-0.25in}
    \caption{\textbf{Effect of student model size on AMD performance.} Accuracy increases with model size, while accuracy gain peaks at 4B.}
    \label{fig:student_size}
    \vspace{-0.2in}
\end{figure}
\autoref{fig:student_size} shows AppWorld accuracy as a function of student model size within the Qwen3 family (1.7B to 14B), using GPT-5-mini as the teacher.
AMD accuracy increases consistently with model size (21.43\%, 49.40\%, 51.79\%, and 52.68\% for 1.7B, 4B, 8B, and 14B, respectively), while the accuracy gain peaks at 4B (+34.52\%p) and diminishes for larger models.
At 1.7B, the student's limited capacity constrains its ability to effectively utilize the injected memories, resulting in modest absolute accuracy despite a non-trivial gain over the zero-shot baseline.
At 8B and 14B, AMD accuracy matches and even slightly surpasses the teacher's performance level (50.00\% on AppWorld), with marginal room for further improvement at this scale.
Together, these results indicate that AMD is effective across a range of student model sizes, with the largest relative benefit accruing to models around 4B that are capable enough to leverage the transferred knowledge while still having substantial room for improvement over their relatively low zero-shot baseline.

\begin{figure}[t!]
    \centering
    \includegraphics[width=\linewidth]{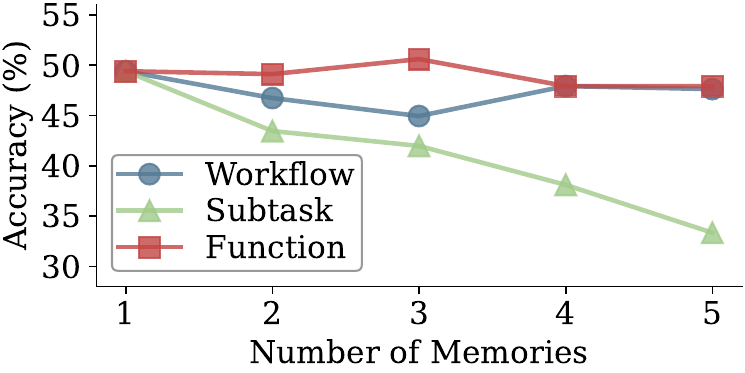}
    \vspace{-0.3in}
    \caption{\textbf{Effect of retrieval count $k$ on AMD performance.} Accuracy at $k{=}1$ is near-optimal for all memory types, and increasing $k$ generally degrades performance.}
    \label{fig:num_memory}
    \vspace{-0.1in}
\end{figure}

\subsection{Effect of Retrieval Count}

In \autoref{fig:num_memory}, we present AppWorld accuracy as the number of retrieved memories per type is varied from 1 to 5 (Qwen3-4B, GPT-5-mini teacher).
Performance at $k{=}1$ is already optimal or near-optimal for all three memory types, and increasing $k$ generally does not improve accuracy.
The effect is most pronounced for \texttt{Subtask} memory, where accuracy drops monotonically from 49.40\% to 33.34\% as $k$ increases.
\texttt{Workflow} memory shows a moderate decline, while \texttt{Function} memory remains stable across $k$.
We attribute this degradation to the limited capacity of small student models: injecting additional lower-ranked memory entries increases the likelihood of introducing irrelevant or loosely matched content, which interferes with task execution.
This highlights the importance of precise, high-confidence memory injection over breadth.

\subsection{Impact of Memory Representation}
We ablate the representation format for each memory type, comparing natural language text against a code-centric format, and report the results in \autoref{tab:memory_format}.
For \texttt{Workflow} memory, a natural language insight outperforms a code-centric format (49.40\% vs.\ 44.05\%), as task-level strategies are more naturally and generalizably expressed in prose.
For \texttt{Subtask} memory, the code-centric format, which pairs a brief textual description with concrete executable code blocks, outperforms a purely natural language description (49.40\% vs.\ 23.21\%). 
Executable examples provide unambiguous, directly actionable patterns for API invocation that small models can follow more reliably than abstract descriptions.
Replacing \texttt{Function} memory with natural language also degrades performance (47.62\% vs.\ 49.40\%), and replacing all three types with natural language yields a large drop (26.19\%), underscoring the importance of retaining concrete code examples at the subtask and function levels.
Together, these results suggest that the optimal representation varies by memory granularity: high-level planning knowledge transfers best as natural language, while low-level execution knowledge is better conveyed through concrete code.

\begin{table}[t]
\centering
\vspace{0.1in}
\caption{\textbf{Effect of memory representation on AppWorld.} Our design choice achieves the best accuracy.}
\label{tab:memory_format}
\vspace{-0.1in}
\setlength{\tabcolsep}{13pt}
\renewcommand{\arraystretch}{0.85}
\resizebox{\linewidth}{!}{%
\begin{tabular}{ccc c}
\toprule
\textbf{Workflow} & \textbf{Sub-task} & \textbf{Function} & \textbf{Acc} \\
\midrule
\midrule
Code & Code & Code & 44.05 \\
Text & Code & Code & \textbf{49.40} \\
Text & Text & Code & 23.21     \\
Text & Code & Text & 47.62 \\
Text & Text & Text & 26.19 \\
\bottomrule
\end{tabular}%
}
\vspace{-0.1in}
\end{table}

\subsection{Qualitative Results}

\autoref{fig:case_study_main} presents a case study on a representative AppWorld task that requires an agent to approve all pending Venmo payment requests received within the current month and subsequently withdraw the remaining balance to a designated card. This task is well-suited for illustrating the cascading contribution of each memory type, as each type resolves a distinct failure mode that the previous configuration could not address.
Without any memory injection, the agent misinterprets the temporal scope of the task and processes requests beyond the current month. \texttt{Workflow} memory corrects this by encoding the implied temporal constraint into the procedural template. This correction, however, exposes a lower-level datetime compatibility failure that traps the agent in a self-repair loop until it exhausts its step budget.
Adding \texttt{Subtask} memory resolves this issue by injecting a verified parsing pattern derived from successful prior trajectories. With the temporal filtering now handled correctly, the agent successfully approves the target requests but encounters a runtime error at the withdrawal stage. The agent's self-repair attempt fails to recover, and the task remains incomplete. 
\texttt{Function} memory resolves this final failure by providing the correct API usage pattern, enabling the agent to extract the balance and complete the full task.
These results demonstrate that the three memory types form a complementary hierarchy operating at distinct layers of agent competence, spanning task planning, execution strategy, and API interaction. All three are necessary for the agent to reach a correct solution on tasks requiring their combined coverage.

\section{Conclusion}
We presented Agent Memory Distillation (AMD), a training-free 
framework for transferring teacher agent experiences to small 
student agents through hierarchically structured memory. We first 
identified that naive memory transfer yields only marginal 
improvements due to the capability gap between teacher and student 
agents, and showed that this gap cannot be bridged by simply 
providing high-quality teacher memories. AMD addresses this by 
constructing three complementary memory types at different levels 
of task granularity: \texttt{Workflow} memory for high-level 
planning, \texttt{Subtask} memory for concrete behavioral 
references, and \texttt{Function} memory for fine-grained tool 
invocation guidance. Experiments across three benchmarks with four 
student models demonstrate that AMD consistently outperforms 
zero-shot baselines and all memory-based baselines, with some 
students approaching or even surpassing teacher-level performance. 
Ablation studies further confirm that each memory type contributes 
distinctly to knowledge transfer, and that effective distillation 
requires careful alignment between memory complexity and student 
capacity. We hope AMD provides a foundation for future research on 
scalable and training-free knowledge transfer for small language 
model agents.

\section*{Limitations}


First, AMD is evaluated only on text-based tool-use benchmarks that involve Python APIs or structured function calls, so its generalization to less structured agentic settings remains unverified. Such settings include multimodal environments in which agents must ground actions in visual observations, as well as coding tasks, where the action space is not a fixed set of callable operations but open-ended code that the agent must generate from scratch. Second, memory is constructed offline from a fixed set of teacher trajectories and remains frozen at inference time. The framework therefore cannot incorporate the student's own successes and failures at test time, nor can it adapt to distribution shifts between the teacher's demonstrations and the tasks the student actually encounters. Finally, AMD depends on the quality and suitability of the teacher trajectories. As shown in Table~\ref{tab:teacher_ablation}, a stronger teacher does not always yield larger student gains, since the benefit also hinges on teacher--student compatibility. Adaptive teacher selection, that is, determining which teacher best suits a given student, thus remains an open problem for future work.



\bibliography{custom}

@article{kim2026memory,
  title={Memory Transfer Learning: How Memories are Transferred Across Domains in Coding Agents},
  author={Kim, Kangsan and Kang, Minki and Kim, Taeil and Yang, Yanlai and Ren, Mengye and Hwang, Sung Ju},
  journal={arXiv preprint arXiv:2604.14004},
  year={2026}
}

@article{ouyang2025reasoningbank,
  title={Reasoningbank: Scaling agent self-evolving with reasoning memory},
  author={Ouyang, Siru and Yan, Jun and Hsu, I and Chen, Yanfei and Jiang, Ke and Wang, Zifeng and Han, Rujun and Le, Long T and Daruki, Samira and Tang, Xiangru and others},
  journal={arXiv preprint arXiv:2509.25140},
  year={2025}
}

@article{wang2024agent,
  title={Agent workflow memory},
  author={Wang, Zora Zhiruo and Mao, Jiayuan and Fried, Daniel and Neubig, Graham},
  journal={arXiv preprint arXiv:2409.07429},
  year={2024}
}

@inproceedings{zhao2024expel,
  title={Expel: Llm agents are experiential learners},
  author={Zhao, Andrew and Huang, Daniel and Xu, Quentin and Lin, Matthieu and Liu, Yong-Jin and Huang, Gao},
  booktitle={Proceedings of the AAAI Conference on Artificial Intelligence},
  volume={38},
  number={17},
  pages={19632--19642},
  year={2024}
}

@article{zhang2026memrl,
  title={Memrl: Self-evolving agents via runtime reinforcement learning on episodic memory},
  author={Zhang, Shengtao and Wang, Jiaqian and Zhou, Ruiwen and Liao, Junwei and Feng, Yuchen and Li, Zhuo and Zheng, Yujie and Zhang, Weinan and Wen, Ying and Li, Zhiyu and others},
  journal={arXiv preprint arXiv:2601.03192},
  year={2026}
}

@article{fang2025memp,
  title={Memp: Exploring agent procedural memory},
  author={Fang, Runnan and Liang, Yuan and Wang, Xiaobin and Wu, Jialong and Qiao, Shuofei and Xie, Pengjun and Huang, Fei and Chen, Huajun and Zhang, Ningyu},
  journal={arXiv preprint arXiv:2508.06433},
  year={2025}
}

@article{xu2026evolution,
  title={The evolution of tool use in llm agents: From single-tool call to multi-tool orchestration},
  author={Xu, Haoyuan and Li, Chang and Ma, Xinyan and Ou, Xianhao and Zhang, Zihan and He, Tao and Liu, Xiangyu and Wang, Zixiang and Liang, Jiafeng and Chu, Zheng and others},
  journal={arXiv preprint arXiv:2603.22862},
  year={2026}
}

@article{du2026memory,
  title={Memory for autonomous llm agents: Mechanisms, evaluation, and emerging frontiers},
  author={Du, Pengfei},
  journal={arXiv preprint arXiv:2603.07670},
  year={2026}
}

@inproceedings{liao2025reflectool,
  title={Reflectool: Towards reflection-aware tool-augmented clinical agents},
  author={Liao, Yusheng and Jiang, Shuyang and Wang, Yanfeng and Wang, Yu},
  booktitle={Proceedings of the 63rd Annual Meeting of the Association for Computational Linguistics (Volume 1: Long Papers)},
  pages={13507--13531},
  year={2025}
}

@article{xia2025experience,
  title={From experience to strategy: Empowering llm agents with trainable graph memory},
  author={Xia, Siyu and Xu, Zekun and Chai, Jiajun and Fan, Wentian and Song, Yan and Wang, Xiaohan and Yin, Guojun and Lin, Wei and Zhang, Haifeng and Wang, Jun},
  journal={arXiv preprint arXiv:2511.07800},
  year={2025}
}

@article{wu2025human,
  title={From human memory to ai memory: A survey on memory mechanisms in the era of llms},
  author={Wu, Yaxiong and Liang, Sheng and Zhang, Chen and Wang, Yichao and Zhang, Yongyue and Guo, Huifeng and Tang, Ruiming and Liu, Yong},
  journal={arXiv preprint arXiv:2504.15965},
  year={2025}
}

@article{luo2026storage,
  title={From Storage to Experience: A Survey on the Evolution of LLM Agent Memory Mechanisms},
  author={Luo, Jinghao and Tian, Yuchen and Cao, Chuxue and Luo, Ziyang and Lin, Hongzhan and Li, Kaixin and Kong, Chuyi and Yang, Ruichao and Ma, Jing},
  year={2026},
  publisher={Preprints}
}

@article{hu2025sample,
  title={Sample-Efficient Online Learning in LM Agents via Hindsight Trajectory Rewriting},
  author={Hu, Michael Y and Van Durme, Benjamin and Andreas, Jacob and Jhamtani, Harsh},
  journal={arXiv preprint arXiv:2510.10304},
  year={2025}
}

@article{allard2026experiential,
  title={Experiential reflective learning for self-improving llm agents},
  author={Allard, Marc-Antoine and Teinturier, Arnaud and Xing, Victor and Viaud, Gautier},
  journal={arXiv preprint arXiv:2603.24639},
  year={2026}
}

@article{ding2026agenther,
  title={AgentHER: Hindsight Experience Replay for LLM Agent Trajectory Relabeling},
  author={Ding, Liang},
  journal={arXiv preprint arXiv:2603.21357},
  year={2026}
}

@article{hinton2015distilling,
  title={Distilling the knowledge in a neural network},
  author={Hinton, Geoffrey and Vinyals, Oriol and Dean, Jeff},
  journal={arXiv preprint arXiv:1503.02531},
  year={2015}
}

@article{kang2026distilling,
  title={Distilling llm agent into small models with retrieval and code tools},
  author={Kang, Minki and Jeong, Jongwon and Lee, Seanie and Cho, Jaewoong and Hwang, Sung Ju},
  journal={Advances in Neural Information Processing Systems},
  volume={38},
  pages={106501--106538},
  year={2026}
}

@inproceedings{mirzadeh2020improved,
  title={Improved knowledge distillation via teacher assistant},
  author={Mirzadeh, Seyed Iman and Farajtabar, Mehrdad and Li, Ang and Levine, Nir and Matsukawa, Akihiro and Ghasemzadeh, Hassan},
  booktitle={Proceedings of the AAAI conference on artificial intelligence},
  volume={34},
  number={04},
  pages={5191--5198},
  year={2020}
}

@article{guo2020reducing,
  title={Reducing the teacher-student gap via spherical knowledge distillation},
  author={Guo, Jia and Chen, Minghao and Hu, Yao and Zhu, Chen and He, Xiaofei and Cai, Deng},
  journal={arXiv preprint arXiv:2010.07485},
  year={2020}
}

@article{wei2022emergent,
  author       = {Jason Wei and
                  Yi Tay and
                  Rishi Bommasani and
                  Colin Raffel and
                  Barret Zoph and
                  Sebastian Borgeaud and
                  Dani Yogatama and
                  Maarten Bosma and
                  Denny Zhou and
                  Donald Metzler and
                  Ed H. Chi and
                  Tatsunori Hashimoto and
                  Oriol Vinyals and
                  Percy Liang and
                  Jeff Dean and
                  William Fedus},
  title        = {Emergent Abilities of Large Language Models},
  journal      = {Trans. Mach. Learn. Res.},
  volume       = {2022},
  year         = {2022},
  url          = {https://openreview.net/forum?id=yzkSU5zdwD},
  bibsource    = {dblp computer science bibliography, https://dblp.org}
}

@article{zhao2024context,
  title={Is in-context learning sufficient for instruction following in llms?},
  author={Zhao, Hao and Andriushchenko, Maksym and Croce, Francesco and Flammarion, Nicolas},
  journal={arXiv preprint arXiv:2405.19874},
  year={2024}
}

@inproceedings{shen2024small,
  title={Small llms are weak tool learners: A multi-llm agent},
  author={Shen, Weizhou and Li, Chenliang and Chen, Hongzhan and Yan, Ming and Quan, Xiaojun and Chen, Hehong and Zhang, Ji and Huang, Fei},
  booktitle={Proceedings of the 2024 conference on empirical methods in natural language processing},
  pages={16658--16680},
  year={2024}
}

@inproceedings{trivedi2024appworld,
  title={Appworld: A controllable world of apps and people for benchmarking interactive coding agents},
  author={Trivedi, Harsh and Khot, Tushar and Hartmann, Mareike and Manku, Ruskin and Dong, Vinty and Li, Edward and Gupta, Shashank and Sabharwal, Ashish and Balasubramanian, Niranjan},
  booktitle={Proceedings of the 62nd Annual Meeting of the Association for Computational Linguistics (Volume 1: Long Papers)},
  pages={16022--16076},
  year={2024}
}

@inproceedings{patil2025bfcl,
    title={The Berkeley Function Calling Leaderboard (BFCL): From Tool Use to Agentic Evaluation of Large Language Models}, 
    author={Patil, Shishir G. and Mao, Huanzhi and Cheng-Jie Ji, Charlie and Yan, Fanjia and Suresh, Vishnu and Stoica, Ion and E. Gonzalez, Joseph},
    booktitle={Forty-second International Conference on Machine Learning},
    year={2025},
}

@misc{lu2024toolsandbox,
      title={ToolSandbox: A Stateful, Conversational, Interactive Evaluation Benchmark for LLM Tool Use Capabilities}, 
      author={Jiarui Lu and Thomas Holleis and Yizhe Zhang and Bernhard Aumayer and Feng Nan and Felix Bai and Shuang Ma and Shen Ma and Mengyu Li and Guoli Yin and Zirui Wang and Ruoming Pang},
      year={2024},
      eprint={2408.04682},
      archivePrefix={arXiv},
      primaryClass={cs.CL},
      url={https://arxiv.org/abs/2408.04682}, 
}

@article{shen2026structurally,
  title={Structurally Aligned Subtask-Level Memory for Software Engineering Agents},
  author={Shen, Kangning and Zhang, Jingyuan and Sun, Chenxi and Zeng, Wencong and Yue, Yang},
  journal={arXiv preprint arXiv:2602.21611},
  year={2026}
}

@article{shinn2023reflexion,
  title={Reflexion: Language agents with verbal reinforcement learning},
  author={Shinn, Noah and Cassano, Federico and Gopinath, Ashwin and Narasimhan, Karthik and Yao, Shunyu},
  journal={Advances in neural information processing systems},
  volume={36},
  pages={8634--8652},
  year={2023}
}

@article{zhang2025memevolve,
  title={Memevolve: Meta-evolution of agent memory systems},
  author={Zhang, Guibin and Ren, Haotian and Zhan, Chong and Zhou, Zhenhong and Wang, Junhao and Zhu, He and Zhou, Wangchunshu and Yan, Shuicheng},
  journal={arXiv preprint arXiv:2512.18746},
  year={2025}
}

@inproceedings{zheng2024synapse,
  title={Synapse: Trajectory-as-exemplar prompting with memory for computer control},
  author={Zheng, Longtao and Wang, Rundong and Wang, Xinrun and An, Bo},
  booktitle={International Conference on Learning Representations},
  volume={2024},
  pages={19036--19066},
  year={2024}
}

@article{tang2025agent,
  title={Agent kb: Leveraging cross-domain experience for agentic problem solving},
  author={Tang, Xiangru and Qin, Tianrui and Peng, Tianhao and Zhou, Ziyang and Shao, Daniel and Du, Tingting and Wei, Xinming and Xia, Peng and Wu, Fang and Zhu, He and others},
  journal={arXiv preprint arXiv:2507.06229},
  year={2025}
}

@article{achiam2023gpt,
  title={Gpt-4 technical report},
  author={Achiam, Josh and Adler, Steven and Agarwal, Sandhini and Ahmad, Lama and Akkaya, Ilge and Aleman, Florencia Leoni and Almeida, Diogo and Altenschmidt, Janko and Altman, Sam and Anadkat, Shyamal and others},
  journal={arXiv preprint arXiv:2303.08774},
  year={2023}
}

@article{comanici2025gemini,
  title={Gemini 2.5: Pushing the frontier with advanced reasoning, multimodality, long context, and next generation agentic capabilities},
  author={Comanici, Gheorghe and Bieber, Eric and Schaekermann, Mike and Pasupat, Ice and Sachdeva, Noveen and Dhillon, Inderjit and Blistein, Marcel and Ram, Ori and Zhang, Dan and Rosen, Evan and others},
  journal={arXiv preprint arXiv:2507.06261},
  year={2025}
}

@inproceedings{gu2024minillm,
  title={Minillm: Knowledge distillation of large language models},
  author={Gu, Yuxian and Dong, Li and Wei, Furu and Huang, Minlie},
  booktitle={International Conference on Learning Representations},
  volume={2024},
  pages={32694--32717},
  year={2024}
}

@article{guo2025deepseek,
  title={Deepseek-r1: Incentivizing reasoning capability in llms via reinforcement learning},
  author={Guo, Daya and Yang, Dejian and Zhang, Haowei and Song, Junxiao and Wang, Peiyi and Zhu, Qihao and Xu, Runxin and Zhang, Ruoyu and Ma, Shirong and Bi, Xiao and others},
  journal={arXiv preprint arXiv:2501.12948},
  year={2025}
}

@inproceedings{hsieh2023distilling,
  title={Distilling step-by-step! outperforming larger language models with less training data and smaller model sizes},
  author={Hsieh, Cheng-Yu and Li, Chun-Liang and Yeh, Chih-Kuan and Nakhost, Hootan and Fujii, Yasuhisa and Ratner, Alex and Krishna, Ranjay and Lee, Chen-Yu and Pfister, Tomas},
  booktitle={Findings of the Association for Computational Linguistics: ACL 2023},
  pages={8003--8017},
  year={2023}
}

@article{sanh2019distilbert,
  title={DistilBERT, a distilled version of BERT: smaller, faster, cheaper and lighter},
  author={Sanh, Victor and Debut, Lysandre and Chaumond, Julien and Wolf, Thomas},
  journal={arXiv preprint arXiv:1910.01108},
  year={2019}
}

@inproceedings{xu2025speculative,
  title={Speculative knowledge distillation: Bridging the teacher-student gap through interleaved sampling},
  author={Xu, Wenda and Han, Rujun and Wang, Zifeng and Le, Long and Madeka, Dhruv and Li, Lei and Wang, William and Agarwal, Rishabh and Lee, Chen-Yu and Pfister, Tomas},
  booktitle={International Conference on Learning Representations},
  volume={2025},
  pages={64616--64646},
  year={2025}
}

@article{liu2025structured,
  title={Structured agent distillation for large language model},
  author={Liu, Jun and Kong, Zhenglun and Dong, Peiyan and Yang, Changdi and Li, Tianqi and Tang, Hao and Yuan, Geng and Niu, Wei and Zhang, Wenbin and Zhao, Pu and others},
  journal={arXiv preprint arXiv:2505.13820},
  year={2025}
}

@article{lyu2025correction,
  title={From correction to mastery: Reinforced distillation of large language model agents},
  author={Lyu, Yuanjie and Wang, Chengyu and Huang, Jun and Xu, Tong},
  journal={arXiv preprint arXiv:2509.14257},
  year={2025}
}

@article{qiu2025agentdistill,
  title={Agentdistill: Training-free agent distillation with generalizable mcp boxes},
  author={Qiu, Jiahao and Juan, Xinzhe and Wang, Yimin and Yang, Ling and Qi, Xuan and Zhang, Tongcheng and Guo, Jiacheng and Lu, Yifu and Yao, Zixin and Wang, Hongru and others},
  journal={arXiv preprint arXiv:2506.14728},
  year={2025}
}

@article{yang2025qwen3,
  title={Qwen3 technical report},
  author={Yang, An and Li, Anfeng and Yang, Baosong and Zhang, Beichen and Hui, Binyuan and Zheng, Bo and Yu, Bowen and Gao, Chang and Huang, Chengen and Lv, Chenxu and others},
  journal={arXiv preprint arXiv:2505.09388},
  year={2025}
}

@misc{gemma4_e4b_2026,
  author       = {{Google DeepMind}},
  title        = {Gemma 4 E4B Instruct},
  year         = {2026},
  howpublished = {\url{https://huggingface.co/google/gemma-4-E4B-it}},
  note         = {Accessed: 2026-05-25}
}

@article{grattafiori2024llama,
  title={The llama 3 herd of models},
  author={Grattafiori, Aaron and Dubey, Abhimanyu and Jauhri, Abhinav and Pandey, Abhinav and Kadian, Abhishek and Al-Dahle, Ahmad and Letman, Aiesha and Mathur, Akhil and Schelten, Alan and Vaughan, Alex and others},
  journal={arXiv preprint arXiv:2407.21783},
  year={2024}
}

@article{ibrahim2025finetuning,
  title={Fine-tuning with RAG for Improving LLM Learning of New Skills},
  author={Ibrahim, Humaid and Rozanov, Nikolai and Rei, Marek},
  journal={arXiv preprint arXiv:2510.01375},
  year={2025}
}

\clearpage
\appendix
\providecommand{\algorithmautorefname}{Algorithm}

\section{Algorithmic Details}
\autoref{alg:amd_inference} presents the complete AMD inference pipeline.
At the beginning of each task, the student agent $\pi^S$ retrieves the top-$k$
workflow memory entry from $\mathcal{M}^{wf}$ using the task instruction $s$ as
the retrieval query, and the retrieved insight is prepended to the system prompt.
$\pi^S$ then decomposes $s$ into an ordered sequence of subtask labels
$\mathcal{T} = \{t_1, \ldots, t_L\}$, each of which is used as an independent
retrieval query against $\mathcal{M}^{st}$. Deduplication is applied across
retrieved segments to prevent the same segment from being injected multiple times
when multiple subtask labels map to the same memory entry. Both the workflow
insight and the retrieved subtask segments are injected into the system prompt
before any tool calls are issued.

During the execution loop, each tool call $a_i \in \mathcal{F}$  is conditioned on the preceding observation $o_{i-1}$. When $o_{i-1}$ reports an error, the failing function names are extracted from it. The extracted names serve as the query for looking up candidate records in $\mathcal{M}^{fn}$. The retrieved records are then added to the context used to generate $a_i$. When multiple records exist for the same function, they are ranked by the cosine similarity between the current task context, and the top-$k$ examples are formatted as a hint block. This reactive injection is applied only upon failure, so that the context is not inflated during successful execution. The loop continues until the task is marked complete or the maximum number of steps $T_{\max}$ is reached.

\begin{algorithm*}[t]
    \centering
    \caption{AMD: Student Agent Inference with Hierarchical Memory}
    \label{alg:amd_inference}
    \includegraphics[width=\textwidth]{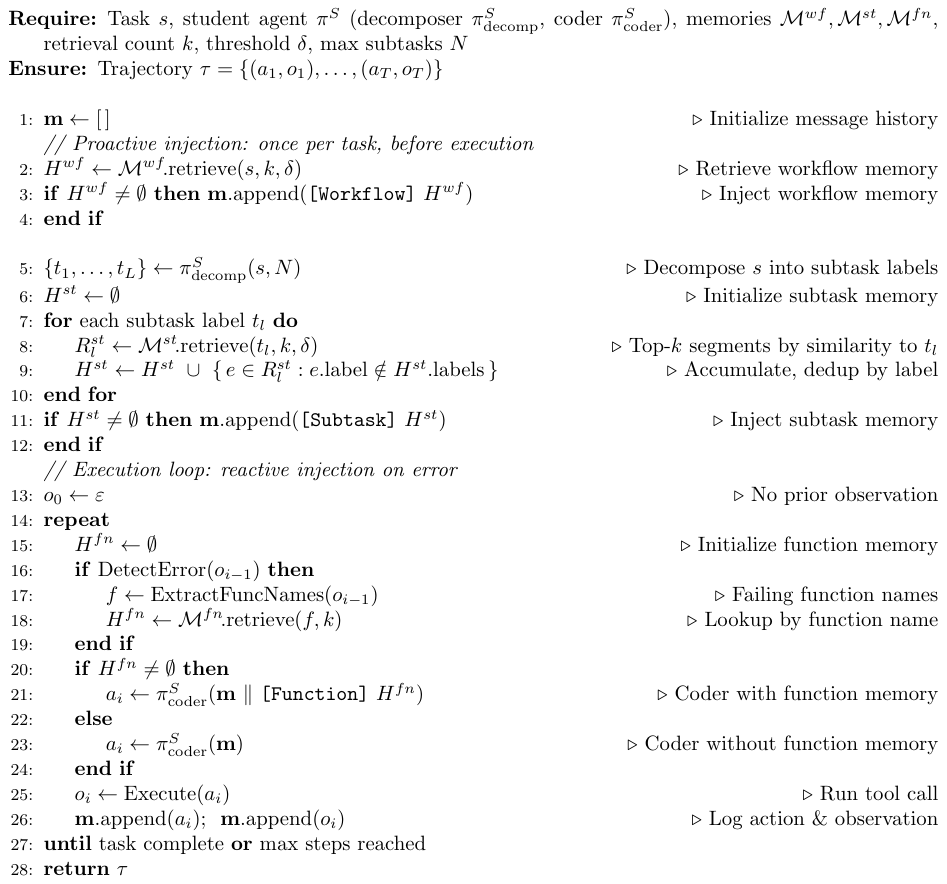}
\end{algorithm*}

\section{Additional Experiment Results}

\subsection{Impact of Each Memory Type}
\autoref{tab:method_ablation} presents the full ablation results across all four
student models on AppWorld and BFCL V3. The contribution ordering holds for almost every model and benchmark. Subtask (ST) memory provides the largest incremental gain when added on top of Workflow (WF) memory, and Function (FN) memory contributes additional but smaller improvements.




Several model-specific patterns are worth noting. For LLaMA3.1-8B on AppWorld,
adding FN memory to the WF+ST configuration decreases accuracy from $30.36\%$ to
$27.38\%$, a trend not observed in the other models. We attribute this to the
relatively weaker instruction-following capacity of LLaMA3.1-8B. The additional
context introduced by Function memory during error recovery may cause the model to
deviate from the planned execution rather than correct its behavior. This
observation is consistent with the general finding that weaker models benefit less
from additional memory injection when their comprehension capacity is already near
its limit.

For Gemma4-E4B, adding FN memory consistently improves performance across both
benchmarks and all configurations. On AppWorld, for instance, accuracy rises from
$30.36\%$ under WF to $40.48\%$ under WF+FN. This suggests that the model has a
relatively stronger capacity for incorporating reactive error-correction guidance
than models of similar scale.

Across all models, the Student Memory variant yields results close to zero-shot on
AppWorld and substantially below AMD on both benchmarks. This confirms that the
quality of teacher trajectories is a critical factor. The student's own experience
is dominated by unsuccessful trajectories and therefore does not provide the
structured and reliable knowledge necessary for effective memory-guided execution.

\subsection{Detailed Case Studies by Memory Type}
\begin{figure*}[h]
    \centering
    \vspace{-0.1in}
    \includegraphics[width=\textwidth]{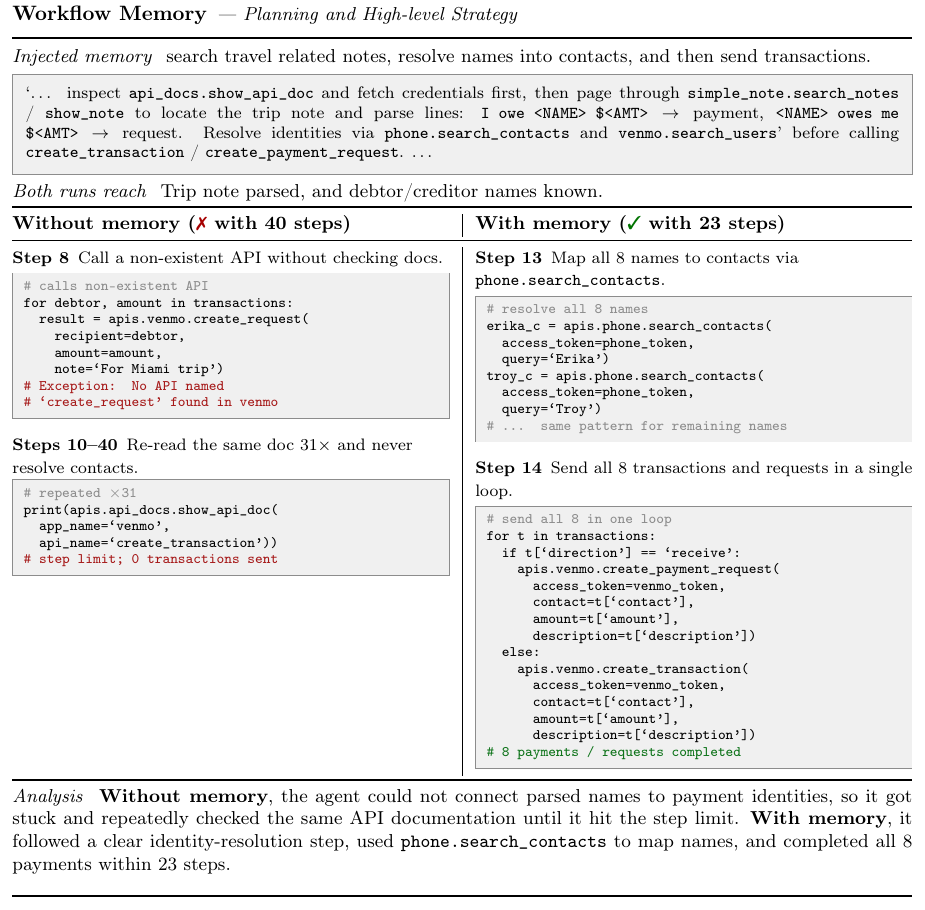}
    \caption{Case Study for Workflow Memory}
    \label{fig:case_study_wf_v2}
    \vspace{-0.15in}
\end{figure*}

\begin{figure*}[h]
    \centering
    \vspace{-0.1in}
    \includegraphics[width=\textwidth]{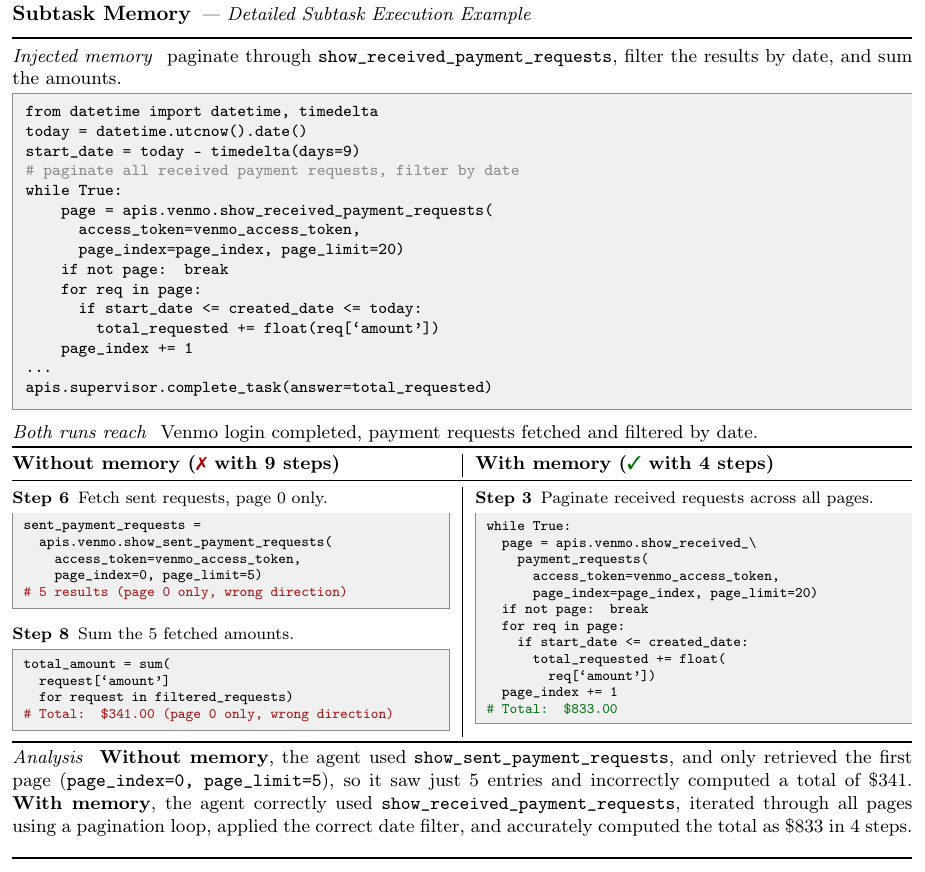}
    \caption{Case Study for Subtask Memory}
    \label{fig:case_study_st_v2}
    \vspace{-0.15in}
\end{figure*}

\begin{figure*}[h]
    \centering
    \vspace{-0.1in}
    \includegraphics[width=\textwidth]{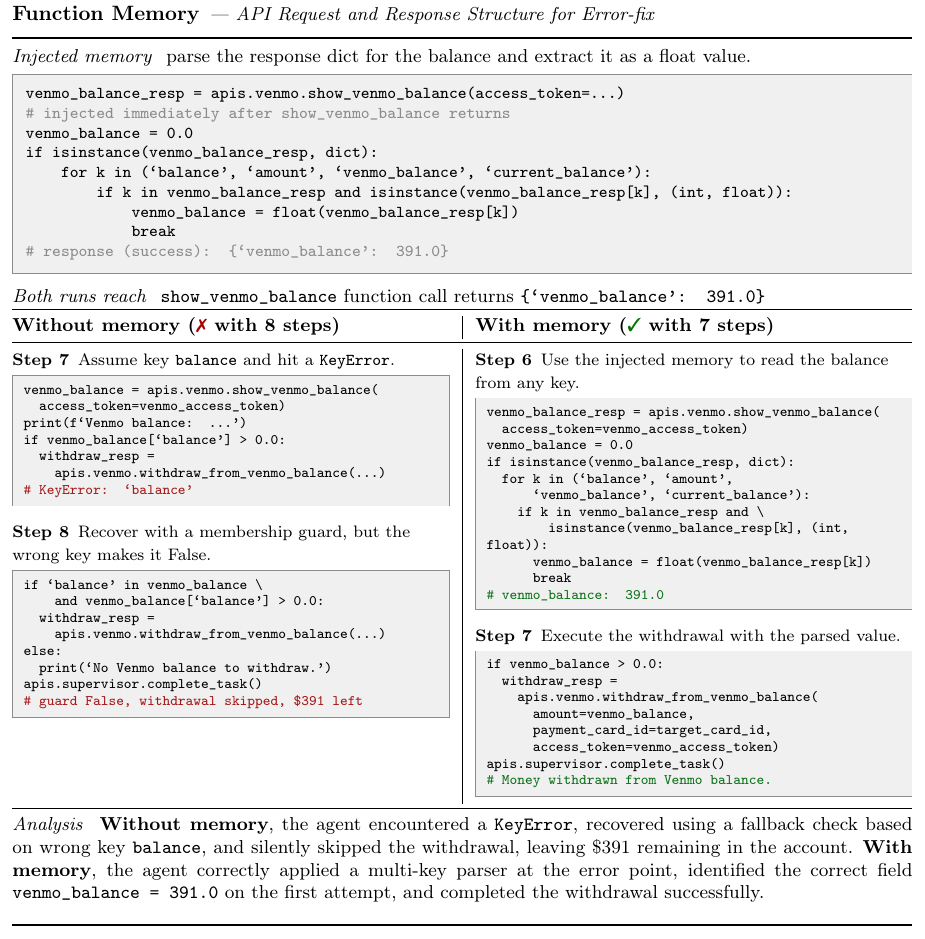}
    \caption{Case Study for Function Memory}
    \label{fig:case_study_fn_v2}
    \vspace{-0.15in}
\end{figure*}

\autoref{fig:case_study_wf_v2}, \autoref{fig:case_study_st_v2}, \autoref{fig:case_study_fn_v2} present three representative case studies corresponding to workflow, subtask, and function memory, respectively, illustrating how injected memory alters the agent's behavior. In each case, the agent without memory and the agent with memory both reach the same intermediate state, but only the memory-augmented agent selects the correct next action. Without memory, the Qwen3-4B agent receives only the task instruction and lacks the procedural knowledge to act on it. These cases show that each failure is resolved when the teacher-generated memory provides the specific procedural knowledge that the small model is missing, such as a planning step, an API call sequence, or a response structure.

\paragraph{Workflow Memory.} Workflow memory captures high-level planning strategies and it is injected upfront before execution begins. In \autoref{fig:case_study_wf_v2}, the agent without memory parses the trip note but does not know the intermediate step of resolving names into contact identities, so it calls a non-existent API and then repeatedly revisits the same API documentation until reaching the step limit. Injecting the workflow memory supplies the identity resolution procedure, enabling the agent to map all names and complete the transactions in substantially fewer steps.

\paragraph{Subtask Memory.} Subtask memory records correct API call sequences for sub-goals and it is also injected upfront. In \autoref{fig:case_study_st_v2}, the agent without memory calls the wrong API endpoint and retrieves only a single page, producing an incorrect total of \$341 instead of the correct \$833. The subtask memory identifies the correct API call with a pagination loop, allowing the agent to retrieve all pages and compute the accurate total.

\paragraph{Function Memory.} Function memory provides a successful API call example from the teacher when the student's API call fails, serving as an on-demand correction at the point of failure. In \autoref{fig:case_study_fn_v2}, the agent without memory parses the API response with the wrong key and silently skips a \$391 withdrawal. The function memory provides a multi-key parser derived from the teacher's successful call, so the agent reads the correct response field on the first attempt and completes the withdrawal.

\subsection{Robustness of Distilled Memory under Disjoint Evaluation}
\label{app:disjoint}

In our main experiments, memory is distilled and applied within the same benchmark.
This raises a natural question about whether the observed gains reflect the real
utility of the distilled memory, or whether a task could simply be benefiting from
memory derived from the same task. To answer this, we evaluate our method under
two protocols in which the tasks used to build memory are kept separate from those
used at evaluation, so that every gain comes only from memory distilled from other
tasks. All experiments in this section use Qwen3-4B as the student model.

\paragraph{Cross-split evaluation}
We split each benchmark into disjoint memory-construction and evaluation subsets at a $7{:}3$ ratio,
distill memory from the memory-construction subset alone, and test on the held-out evaluation
subset.

\paragraph{Self-excluded retrieval}
We distill memory over the full benchmark, but during evaluation, each task retrieves memory distilled from other tasks, never its own. Compared to the cross-split setup, this preserves a larger memory pool while ensuring that no task uses memory distilled from itself.

\paragraph{Results}
\autoref{tab:disjoint} reports success rates for both protocols on AppWorld,
BFCL V3, and ToolSandbox. Both protocols preserve
monotonic gains from Zero-shot to WF, WF+ST, and WF+ST+FN, and the full
configuration clearly outperforms the Zero-shot baseline on every benchmark. Since
no task can rely on memory from itself in either setup, these results indicate that
the improvements come from the utility of the distilled memory rather than from
memory construction and evaluation sharing the same tasks. On AppWorld,
self-excluded retrieval slightly outperforms cross-split evaluation,
which is consistent with the larger memory pool available when distilling over the
full task set. We further note that the full configuration (WF+ST+FN) results under both
protocols stay close to the final AMD accuracy for Qwen3-4B in the main results
(\autoref{tab:main_acc}), indicating that AMD retains substantial gains even when memory construction and evaluation are disjoint.

\subsection{Robustness of Distilled Memory across Repeated Runs}
\label{app:repeated_runs}

We assess run-to-run robustness by repeating the zero-shot and AMD configurations five times under the same experimental setup, using Qwen3-4B as the student model. \autoref{tab:repeated_runs} reports the mean and standard deviation of the success rate across the five runs. AMD achieves higher mean success rates than the zero-shot baseline across all three benchmarks. The standard deviations remain below one percentage point in all settings and are substantially smaller than the corresponding mean improvements, demonstrating that AMD's gains are stable across repeated runs.

\begin{table*}[t]
\centering
\small
\vspace{-0.1in}
\caption{Mean $\pm$ standard deviation of success rate (\%) across five runs for each benchmark with Qwen3-4B as the student model.}
\label{tab:repeated_runs}
\vspace{-0.1in}
\begin{tabular}{lccc}
\toprule
\textbf{Method} & \textbf{AppWorld} & \textbf{BFCL V3} & \textbf{ToolSandbox} \\
\midrule
Zero-shot & $14.48 \pm 0.69$ & $15.33 \pm 0.76$ & $16.54 \pm 0.44$ \\
AMD       & $\mathbf{49.60 \pm 0.91}$ & $\mathbf{38.67 \pm 0.58}$ & $\mathbf{20.16 \pm 0.78}$ \\
\bottomrule
\end{tabular}
\end{table*}

\begin{table*}[t]
\centering
\small
\setlength{\tabcolsep}{6pt}
\renewcommand{\arraystretch}{1.1}
\caption{Success rates under the two disjoint-evaluation protocols with Qwen3-4B as the student model. WF, ST, and FN
denote workflow, subtask, and function memory.}
\label{tab:disjoint}
\vspace{-0.1in}
\begin{tabular}{llccc}
\toprule
\multirow{2}{*}{\textbf{Protocol}} & \multirow{2}{*}{\textbf{Memory}}
 & \multicolumn{3}{c}{\textbf{Success Rate (\%)}} \\
\cmidrule(lr){3-5}
 & & \textbf{AppWorld} & \textbf{BFCL V3} & \textbf{ToolSandbox} \\
\midrule
\multirow{4}{*}{Cross-split evaluation}
 & Zero-shot  & 16.07 & 16.13 & 16.67 \\
 & WF         & 21.43 & 25.81 & 18.75 \\
 & WF + ST      & 39.29 & 27.42 & 20.83 \\
 & WF + ST + FN   & \textbf{41.07} & \textbf{30.65} & \textbf{22.92} \\
\midrule
\multirow{4}{*}{Self-excluded retrieval}
 & Zero-shot  & 14.88 & 15.50 & 16.28 \\
 & WF         & 23.21 & 22.50 & 19.38 \\
 & WF + ST      & 40.48 & 30.50 & 20.16 \\
 & WF + ST + FN   & \textbf{46.43} & \textbf{31.00} & \textbf{23.26} \\
\bottomrule
\end{tabular}
\end{table*}

\section{Additional Implementation Details}

\paragraph{Memory Generation}
For \texttt{Workflow} memory generation, we use the prompt shown in \autoref{fig:wf_prompt}.
The teacher LLM is instructed to produce a single short insight describing the overall task-completion strategy at a high level, covering the apps and tools involved, key preconditions, decision rules, validation cues, and common failure patterns to avoid, and explicitly mentioning key API functions in \texttt{app.function\_name} form.
Concrete runtime values such as user IDs, credentials, and file paths are replaced with typed placeholders (e.g., \texttt{<ID>}, \texttt{<EMAIL>}, \texttt{<FILE\_PATH>}), so that the insight remains applicable to future tasks with different specific inputs.
For \texttt{Subtask} memory segmentation, we use the prompt shown in \autoref{fig:st_prompt}.
The teacher LLM is prompted to identify semantic boundaries within the trajectory and to produce, for each resulting segment, a short phrase label and a one-sentence description.
To guide this process, we additionally provide the teacher LLM with rule-based breakpoint hints derived from API call boundaries, and the prompt recommends keeping the number of subtask segments per trajectory to at most six.
For \texttt{Function} memory, each memory stores the function name together with a concrete example drawn from a successful teacher trajectory.
This example comprises the function call and its surrounding context, from which the rationale for the call can be inferred.
For benchmarks with rich API documentation such as AppWorld, we additionally attach the corresponding argument and response schema to each memory.
For benchmarks with minimal tool schemas, such as BFCL V3 and ToolSandbox, each memory uses only the concrete example drawn from the successful teacher trajectory.

\paragraph{Memory Retrieval}
All memory entries are encoded using OpenAI's \texttt{text-embedding-3-small}
model. Retrieval is performed via cosine similarity between the query embedding
and all stored memory embeddings. A minimum similarity threshold $\delta$ is
applied, below which candidates are discarded. We set $k{=}1$ for all three
memory types in the main experiments, retrieving the top-1 Workflow entry, the
top-1 Subtask segment per decomposed subtask label, and the top-1 Function record
per failing function. Deduplication is applied across retrieved Subtask segments
to prevent the same segment from being injected multiple times when multiple
subtask labels map to the same memory entry.

\paragraph{Student Subtask Decomposition}
At the beginning of each task, $\pi^S$ is prompted to decompose the task
instruction into an ordered sequence of subtask labels using the prompt shown in
\autoref{fig:st_decomp_prompt}. Decomposition is performed prior to any tool
calls and is constrained to at most six subtasks. The resulting labels are then
used as independent retrieval queries against $\mathcal{M}^{st}$.


\paragraph{Benchmark-Specific Details}
On AppWorld, the teacher agent accesses Python API calls and a supervisor application for retrieving credentials, and we evaluate on the test-normal subset of 168 tasks. Since AppWorld provides rich API documentation, each function memory is augmented with the corresponding argument and response schema.
On BFCL V3, we use the multi-turn base subset of 200 tasks, and on ToolSandbox we use the base subset of 129 tasks.
Both benchmarks have only minimal, signature-derived tool schemas rather than rich documentation, so on both we build function memory solely from the concrete examples in successful teacher trajectories.

\paragraph{Inference}
All student agents are allowed a maximum of 40 interaction steps per task and are served with vLLM for efficient inference. For the Qwen3 models, we disable thinking mode during inference to ensure a fair comparison with the other student models. The teacher agent, GPT-5-mini, is queried through the OpenAI API.


\section{Memory Examples and Generation Prompts}
\begin{figure*}[h]
    \centering
    \vspace{-0.1in}
    \includegraphics[width=\textwidth]{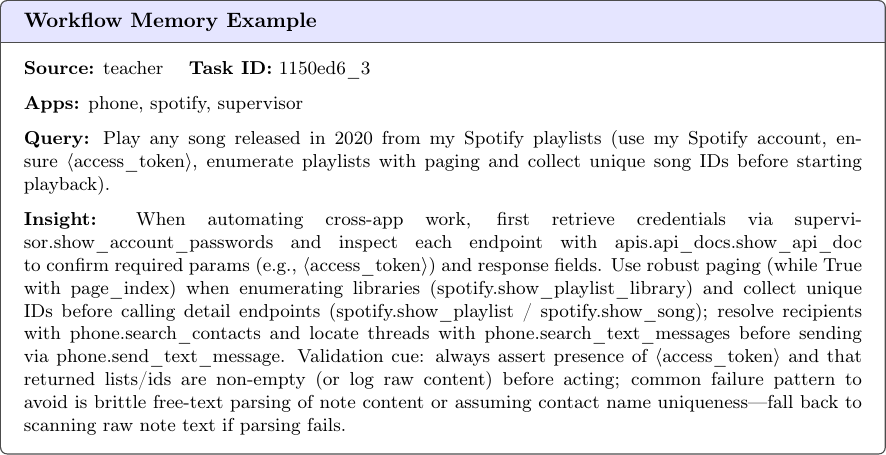}
    \caption{Workflow Memory Example}
    \label{fig:mem_example_start}
    \vspace{-0.15in}
\end{figure*}

\begin{figure*}[h]
    \centering
    \vspace{-0.1in}
    \includegraphics[width=\textwidth]{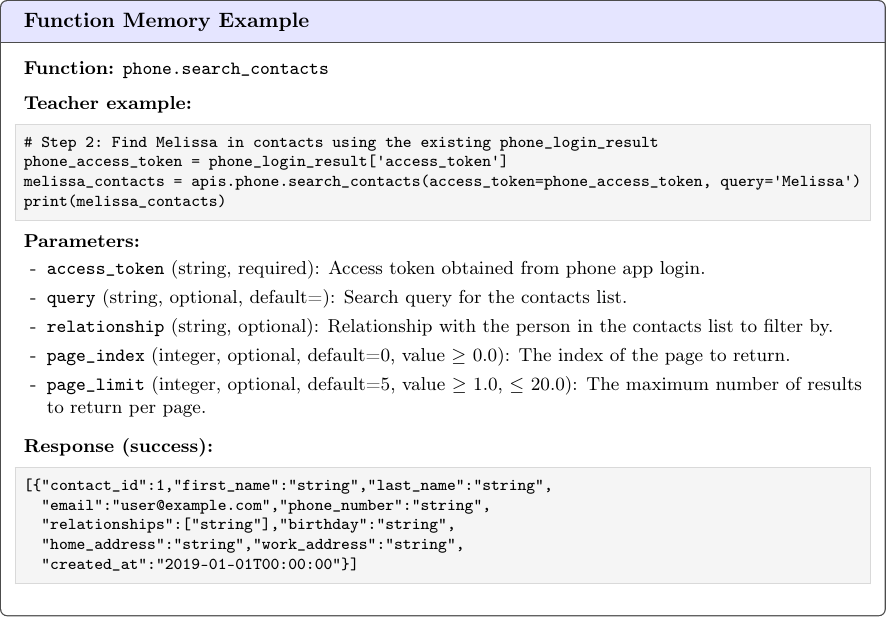}
    \caption{Function Memory Example}
    \label{fig:mem_example_end}
    \vspace{-0.15in}
\end{figure*}

\begin{figure*}[h]
    \centering
    \vspace{-0.1in}
    \includegraphics[width=\textwidth]{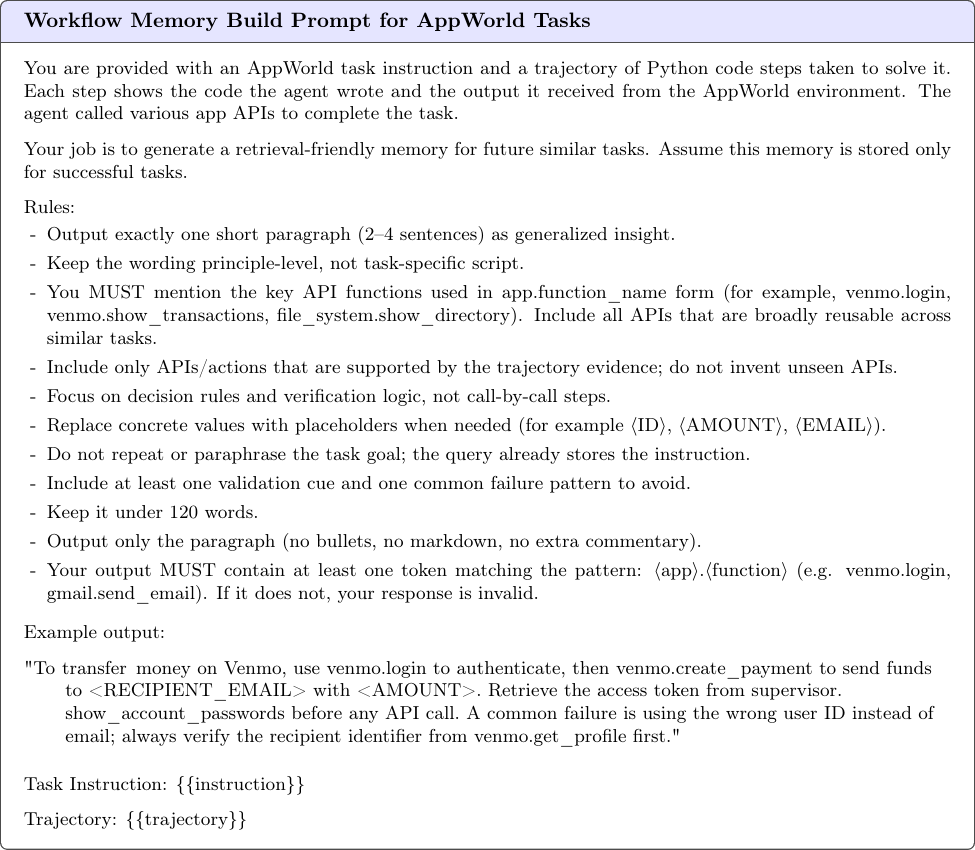}
    \caption{Workflow Memory Build Prompt for AppWorld Tasks}
    \label{fig:wf_prompt}
    \vspace{-0.15in}
\end{figure*}

\begin{figure*}[h]
    \centering
    \vspace{-0.1in}
    \includegraphics[width=\textwidth]{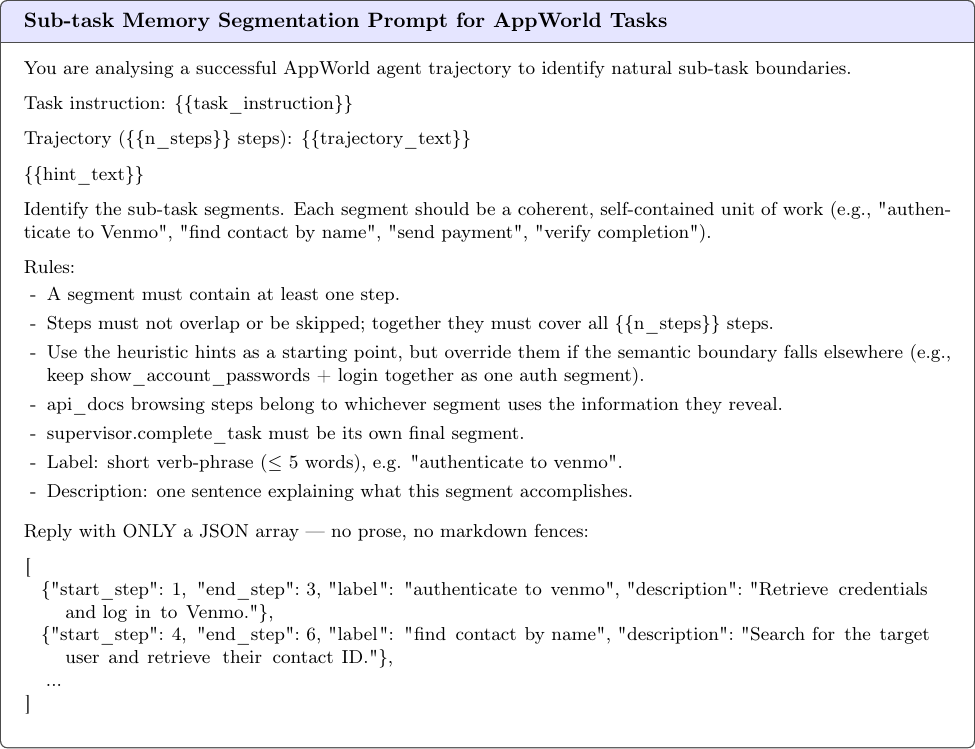}
    \caption{Sub-task Memory Segmentation Prompt for AppWorld Tasks}
    \label{fig:st_prompt}
    \vspace{-0.15in}
\end{figure*}

\begin{figure*}[h]
    \centering
    \vspace{-0.1in}
    \includegraphics[width=\textwidth]{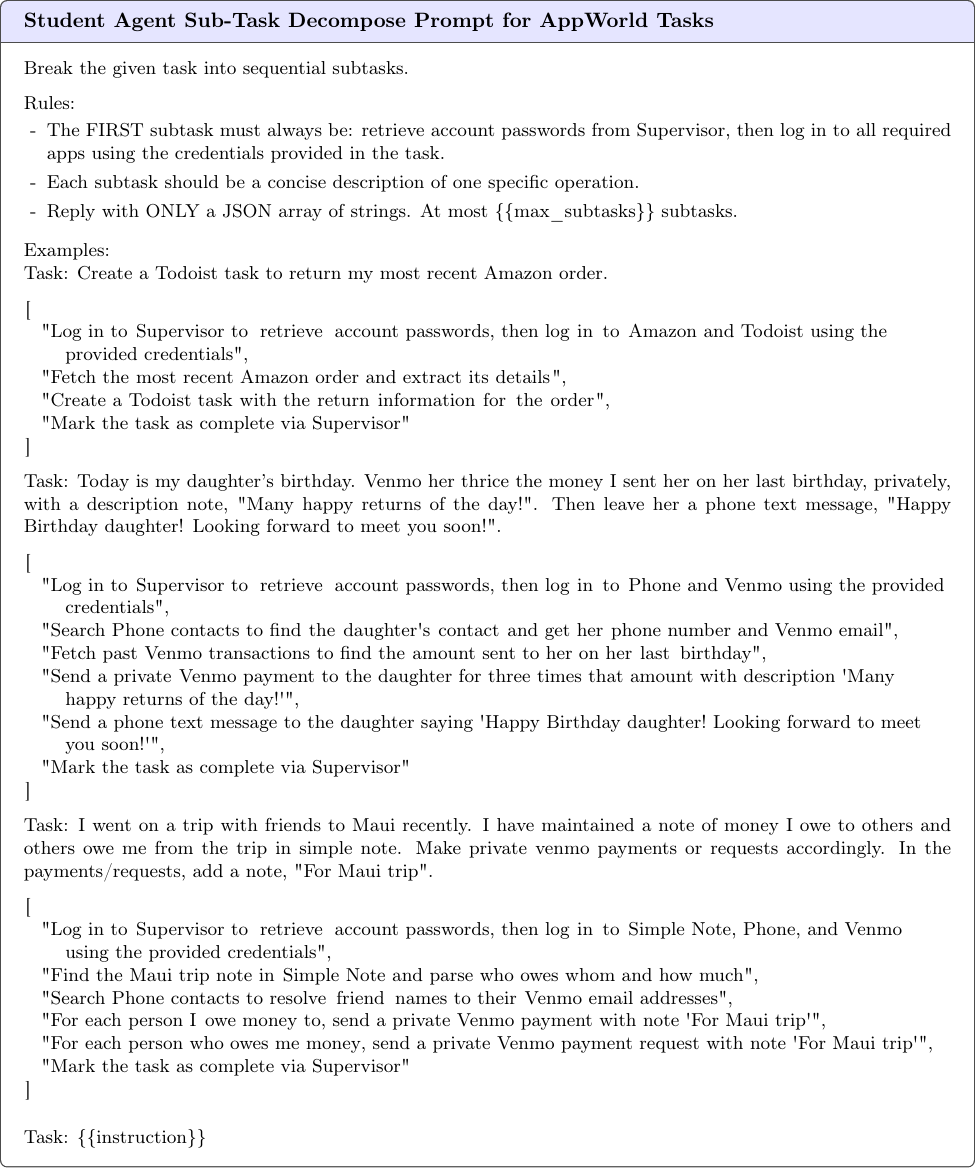}
    \caption{Student Agent Sub-task Decompose Prompt for AppWorld Tasks}
    \label{fig:st_decomp_prompt}
    \vspace{-0.15in}
\end{figure*}

We present concrete examples and generation details for each of the three memory types introduced in our framework. \autoref{fig:mem_example_start} through \autoref{fig:mem_example_end} show representative Workflow, Subtask, and Function memory entries, illustrating how successful teacher trajectories are distilled into structured, reusable knowledge at each level of granularity. We also provide the prompts used to construct these memories, namely the Workflow generation prompt in \autoref{fig:wf_prompt} and the Subtask segmentation prompt in \autoref{fig:st_prompt}. Finally, \autoref{fig:st_decomp_prompt} shows the prompt with which the student agent decomposes each task instruction into an ordered list of subtask labels, which then serve as queries for retrieving Subtask memory.




\begin{figure*}[t!p]
    \centering
    \includegraphics[width=\linewidth]{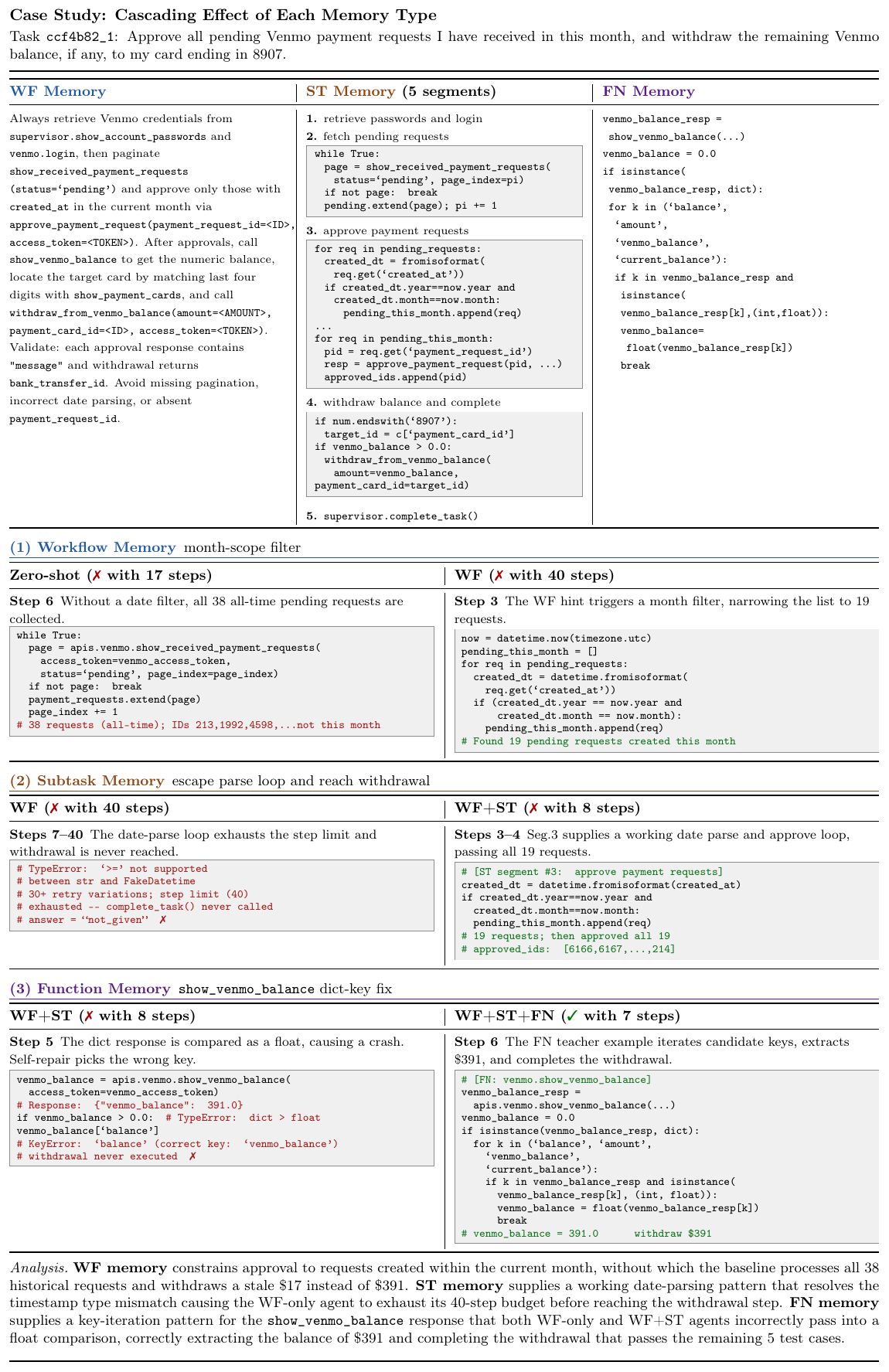}
    \vspace{-0.25in}
    \caption{Cascading Effect of Each Memory Type}
    \label{fig:case_study_main}
    \vspace{0.1in}
\end{figure*}

\begin{figure*}[h]
    \centering
    \vspace{-0.1in}
    \includegraphics[width=\textwidth]{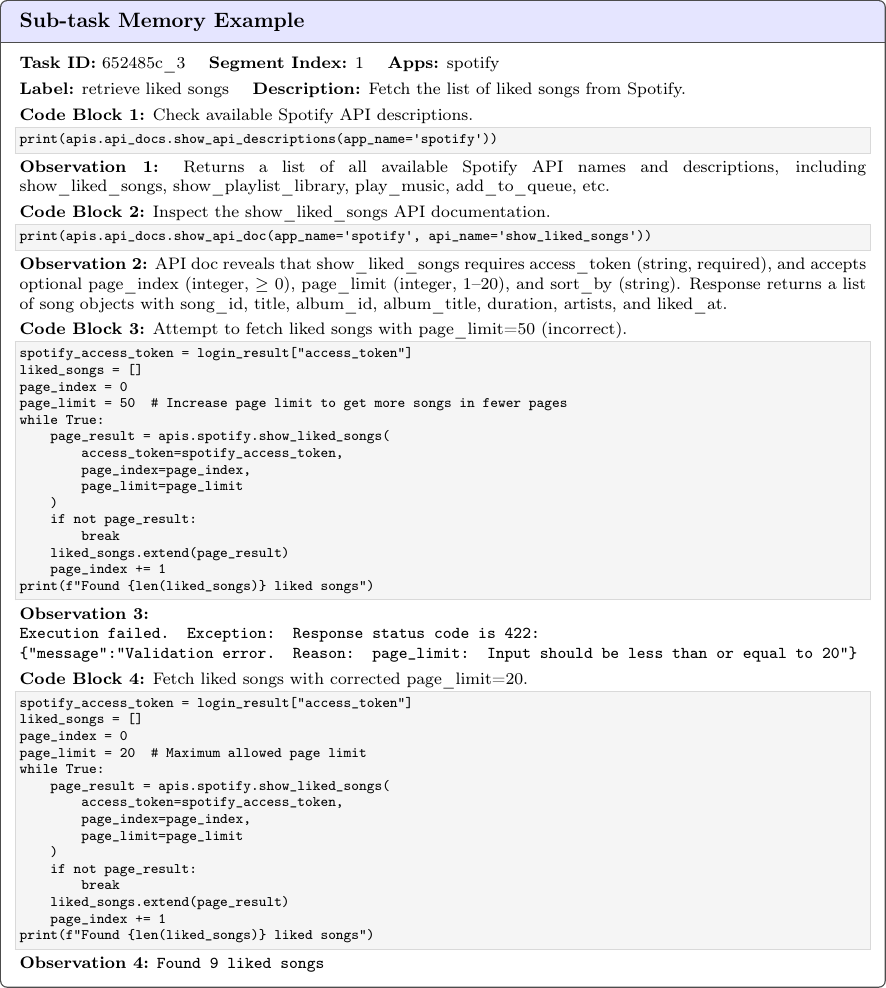}
    \caption{Sub-task Memory Example}
    \vspace{-0.15in}
\end{figure*}

 \end{document}